\documentclass[12pt]{article}
\usepackage{amsmath, amsfonts}
\usepackage{graphicx,psfrag,epsf}
\usepackage{enumerate}
\usepackage{natbib}
\usepackage{xcolor}
\usepackage{soul}
\usepackage{float}
\usepackage[hidelinks]{hyperref}
\usepackage{bm}
\usepackage{graphicx}
\newcommand{\blind}{0}

\begin{document}

\def\spacingset#1{\renewcommand{\baselinestretch}%
{#1}\small\normalsize} \spacingset{1}

%%%%%%%%%%%%%%%%%%%%%%%%%%%%%%%%%%%%%%%%%%%%%%%%%%%%%%%%%%%%%%%%%%%%%%%%%%%%%%

{\if0\blind{
\title{\bf Modeling Spatial Extremes using Non-Gaussian Spatial Autoregressive Models via Convolutional Neural Networks}
\author{
    Sweta Rai, Douglas W.~Nychka, and Soutir Bandyopadhyay
    \vspace{0.5em}
    \thanks{SR, DWN, and SB’s work has been fully supported by the National Science Foundation, CMMI-2210840.\\
    Corresponding author email: sbandyopadhyay@mines.edu}\\
    \hspace{.2cm}Department of Applied Mathematics and Statistics\\ Colorado School of Mines, Golden, CO\\
}
\date{}\maketitle}\else\fi}

\if1\blind
{
  \bigskip
  \bigskip
  \bigskip
  \begin{center}
    {\LARGE\bf Title}
\end{center}
  \medskip
}\fi

\begin{abstract}
\noindent Data derived from remote sensing or numerical simulations often have a regular gridded structure and are large in volume, making it challenging to find accurate spatial models that can fill in missing grid cells or simulate the process effectively, especially in the presence of spatial heterogeneity and heavy-tailed marginal distributions. To overcome this issue, we present a spatial autoregressive modeling framework, which maps observations at a location and its neighbors to independent random variables. This is a highly flexible modeling approach and well-suited for non-Gaussian fields, providing simpler interpretability. In particular, we consider the SAR model with Generalized Extreme Value distribution innovations to combine the observation at a central grid location with its neighbors, capturing extreme spatial behavior based on the heavy-tailed innovations. While these models are fast to simulate by exploiting the sparsity of the key matrices in the computations, the maximum likelihood estimation of the parameters is prohibitive due to the intractability of the likelihood, making optimization challenging. To overcome this, we train a convolutional neural network on a large training set that covers a useful parameter space, and then use the trained network for fast parameter estimation. Finally, we apply this model to analyze annual maximum precipitation data from ERA-Interim-driven Weather Research and Forecasting (WRF) simulations, 
allowing us to explore its spatial extreme behavior across North America.\\

\end{abstract}

\noindent%

\spacingset{1.45}
\newpage
\section{Introduction}
\label{sec:intro}
Large-scale extreme events, such as severe weather, floods, droughts, wildfires, and heatwaves, are expected to become more frequent and more intense due to climate change. Simultaneously, rapid urbanization heightens the need for computationally efficient methods to predict these events in (near) real-time. 
Although the surge in research interest in spatial extremes over the past few decades has led to a well-developed framework for max-stable processes \citep{haan2006extreme, fernandez2003extreme, smith1990max}, there is a growing perspective that strictly enforcing max-stability may not always be adequate for many practical applications \citep{huser2024modeling}, as such models often struggle to capture the localized extremal behavior of several environmental processes \citep{huser2022advances}. In addition, despite the richness of max-stable models, they remain computationally challenging when large numbers of scenarios need to be explored to understand how extreme events may evolve over time. Various approaches have been developed to address these challenges including approximate likelihood methods \citep{huser2014space}, reduced-rank and kernel-based formulations \citep{bopp2021hierarchical, reich2012hierarchical}, and spatial clustering approaches \citep{chen2021assessing}. 

In this work, we propose an alternative to these works, placing an emphasis on computational feasibility for very large non-Gaussian spatial fields while leveraging the well-established framework for Markov random fields (MRF). In particular,  we provide an extension of the Spatial Autoregressive (SAR) framework, traditionally used for Gaussian processes \citep{nychka2015multiresolution, sikorski2024normalizing}, to a non-Gaussian SAR model tailored for spatial extremes. The proposed model shares some tangential similarities with the hierarchical models discussed in \cite{reich2012hierarchical, zhang2023flexible}, in that all these models use basis functions and a nugget term to model spatial dependence in extreme fields. However, the novelty of our approach lies in incorporating the SAR structure within the basis representation, effectively extending the LatticeKrig modeling framework \citep{nychka2015multiresolution} to a non-Gaussian framework by integrating the Generalized Extreme Value (GEV) distribution \citep{jenkinson1955, coles2001introduction}, making the approach suitable for modeling spatial extremes. A key advantage of this new formulation is its computational efficiency, exploiting the sparsity of the key matrices in the construction, making the method practically useful to simulate large-scale spatial extreme events. 

Although our method provides a flexible, computationally efficient framework for simulating and modeling large spatial extreme fields, it also presents a set of complex statistical challenges: (a) only individual components of the model can be interpreted directly, and (b) the model complexity increases with the inclusion of the nugget term. Specifically, with such a complex structure, the likelihood function becomes intractable, and the use of the
maximum likelihood estimation (MLE) is infeasible. Consequently, simulation and artificial intelligence-based inference methods emerge as attractive solutions, providing a highly flexible framework for effectively handling large, non-Gaussian datasets. In recent years, neural network-based approaches for estimating parameters in dependent processes have shown significant promise due to their suitability and computational efficiency in handling complex likelihood structures \citep{gerber2021fast, sainsbury2025neural, richards2024neural, sainsbury2024likelihood, rai2024fast, walchessen2024neural}. These methods exemplify the simulation-based inference paradigm, leveraging a low-cost generative model to construct large training sets of (sample, parameter) pairs. The term `amortized inference' \citep{zammit2024neural} is often used to describe such neural network-based approaches, emphasizing that while training the network is computationally intensive, this cost is offset during deployment, where model parameters can be estimated from samples in a fraction of a second.

Following this, in this work, we train a convolutional neural network (CNN) on samples generated from the extreme SAR model using an intensively designed training set with parameter configurations that cover a wide range of possible scenarios. In our approach, we restrict ourselves to the positive shape GEV distribution (i.e., the Fr\'echet distribution), as it appears to yield desirable properties for modeling spatial extremes. For training the CNN, we fix the input size of the training samples at 16x16 and vary the number of spatial replications of the fields (e.g., 1, 10, and 30 replicates). Although the CNN model can accommodate different input sizes, we prefer a 16×16 window because it facilitates the extraction of small windows from larger spatial domains for evaluation and can also support local likelihood estimation if needed \citep{wiens2020modeling}.

For uncertainty quantification, we propose using quantile regression \citep{koenker2001quantile, koenker2017handbook} to compute the confidence interval (CI). We fit the quantile regression model using the same training data as the CNN model. Although this approach may seem like a natural exercise for statisticians, it appears to be novel in this context, as it has not been previously applied to compute CIs for inference in spatial extremes modeling. Finally, we apply our method to the Regional Climate Model (RCM) precipitation output. We evaluate our model by analyzing annual maximum precipitation derived from ERA-Interim-driven WRF simulations within the NA-CORDEX project \citep{nacordex}, which helps us to explore spatial extremal behavior across North America and to validate the model output against observed process behavior.

The remainder of the paper is structured as follows. In Section \ref{sec:meth}, we discuss the GEV-SAR modeling framework for spatial extremes followed by a description of the neural network used for parameter estimation in Section~\ref{sec:Framework}. Section \ref{sec:sim} presents the results of our simulation study and uncertainty quantification on various test sets to understand the model behavior, along with comparisons to the MLE. Section \ref{sec:CS} discusses the RCM case study and presents diagnostic plots for evaluating the model behavior and findings. Finally, Section \ref{sec:conc} summarizes our findings, examines the strengths and limitations of our model, and presents our conclusions.

\section{The New Modeling Framework For Spatial Extremes}
\label{sec:meth}
This section begins with background information on the GEV distribution, followed by an overview of the general structure of the GEV-SAR model for spatial extremes. Finally, we conclude this section with a thorough explanation of the inference process using the CNN model.

\subsection{Generalized Extreme Value (GEV) Distribution}
In this work, we focus on the GEV distribution as a concrete example of a non-Gaussian modeling framework. However, our approach is flexible and can be extended to accommodate other distributions as well. A GEV distribution with location-scale parameters $\mu\in \mathbb{R}$, $\sigma>0$ and the shape (or tail-index) parameter $\xi\in \mathbb{R}$ has a cumulative distribution  function (cdf)
      \begin{equation}
          F(t) =\left \{ \begin{array}{cc}
          \exp\left[-[1+\xi(t -\mu)/\sigma]_+^{-{1}/{\xi}}\right] &\mathrm{if}\  \xi\neq 0,\\[.02in]
          \exp\left[-\exp( - (t -\mu)/\sigma\right] & \mathrm{if}\ \xi= 0,
          \end{array}
          \right.
          \label{GEV}
      \end{equation}
where for any real number $x$, $x_+=\max\{0,x\}$. We will denote the distribution in (\ref{GEV}) as GEV($\mu, \sigma, \xi$). 
The GEV distribution covers all possible limit distributions of the maxima of  independent and identically distributed (iid) random variables and also for the maxima of dependent random variables satisfying suitable weak dependence conditions \citep{coles2001introduction}. In particular, the GEV distribution takes different forms based on $\xi$: the Gumbel distribution for light-tailed ($\xi = 0$), the Fréchet distribution for heavy-tailed ($\xi > 0$), and the Weibull distribution for short-tailed ($\xi < 0$). The sign of $\xi$ also determines whether the distribution is upper-bounded ($\xi < 0$) or lower-bounded ($\xi > 0$). For this reason we use it as a standard model for the distribution of rare and large events. 

\subsection{The GEV-SAR Model Formulation} Our model integrates the GEV distribution within the SAR framework, creating a novel statistical framework specifically designed for modeling spatial extremes. While similar constructions have been explored in the context of Gaussian processes \citep{nychka2002multiresolution, nychka2015multiresolution}, they remain largely underutilized for modeling
spatial extremes. By leveraging their flexibility and computational efficiency, coupled with modern methods for optimization and inference, this approach holds significant promise for advancing practical applications in modeling spatial extremes. The fundamental concept behind this framework is to represent the process as a weighted sum of basis functions, where the weights themselves are stochastic. A key advantage of this new GEV-SAR formulation is its computational efficiency, exploiting the sparsity of the key matrices in the construction. In this paper, we further improve computational efficiency by incorporating CNN-based inference, as discussed later in Section~\ref{sec:Framework}.

Next, we provide the details of our model. Motivated by the application of a multiplicative framework for modeling spatial extremes as proposed by \cite{reich2012hierarchical}, we assume that spatial observations $y_i$, observed at gridded locations $\bm{s}_i, i = 1, \ldots, n$, follow the structure:
\begin{equation}
\label{obs-model-eq}
y_i = \exp{(\bm{x}_{i}^{T}\bm{d})}\cdot g(\bm{s}_i) \cdot \varepsilon_i,
\end{equation}
where, $\bm{x}_{i}$ is a vector of covariates (e.g., topography, meteorological inputs, land cover, etc.) at the $i$-th location and $\bm{d}$ is a vector of regression coefficients. $g(\bm{s}_i)$ represents the process capturing spatial dependence and variations and $\varepsilon_i$'s, commonly known as the nugget term in spatial statistics, are iid random variables, typically accounting for the measurement error.

\noindent \textbf{Incorporating Covariates in the Model.} For simplicity, the rest of the paper considers model (\ref{obs-model-eq}) without covariates $\bm{x}$. However, we acknowledge that in practice, covariate effects must be accounted for. To do so, one could estimate the regression coefficients using an iterative procedure similar to the backfitting method \citep{hastie1987generalized}. Specifically, one could begin with initial values of $\bm{d}$ and use (\ref{obs-model-eq}), which yields $\log{y_{i}} = \bm{x}_{i}^{T}\bm{d} + \log{(g(\bm{s}_i) \cdot \varepsilon_i)} = \mathrm{I} + \mathrm{II}$ (say).  The parameters in part II can then be estimated using the method described in Section~\ref{sec:Framework}, and these estimates can be used to update $\bm{d}$. These steps could be iterated until convergence is achieved. Furthermore, resampling methods could be used to quantify the uncertainty of the estimated fixed parameters.

\begin{figure}[ht]
\centering
\includegraphics[height = 3in, width=\textwidth]{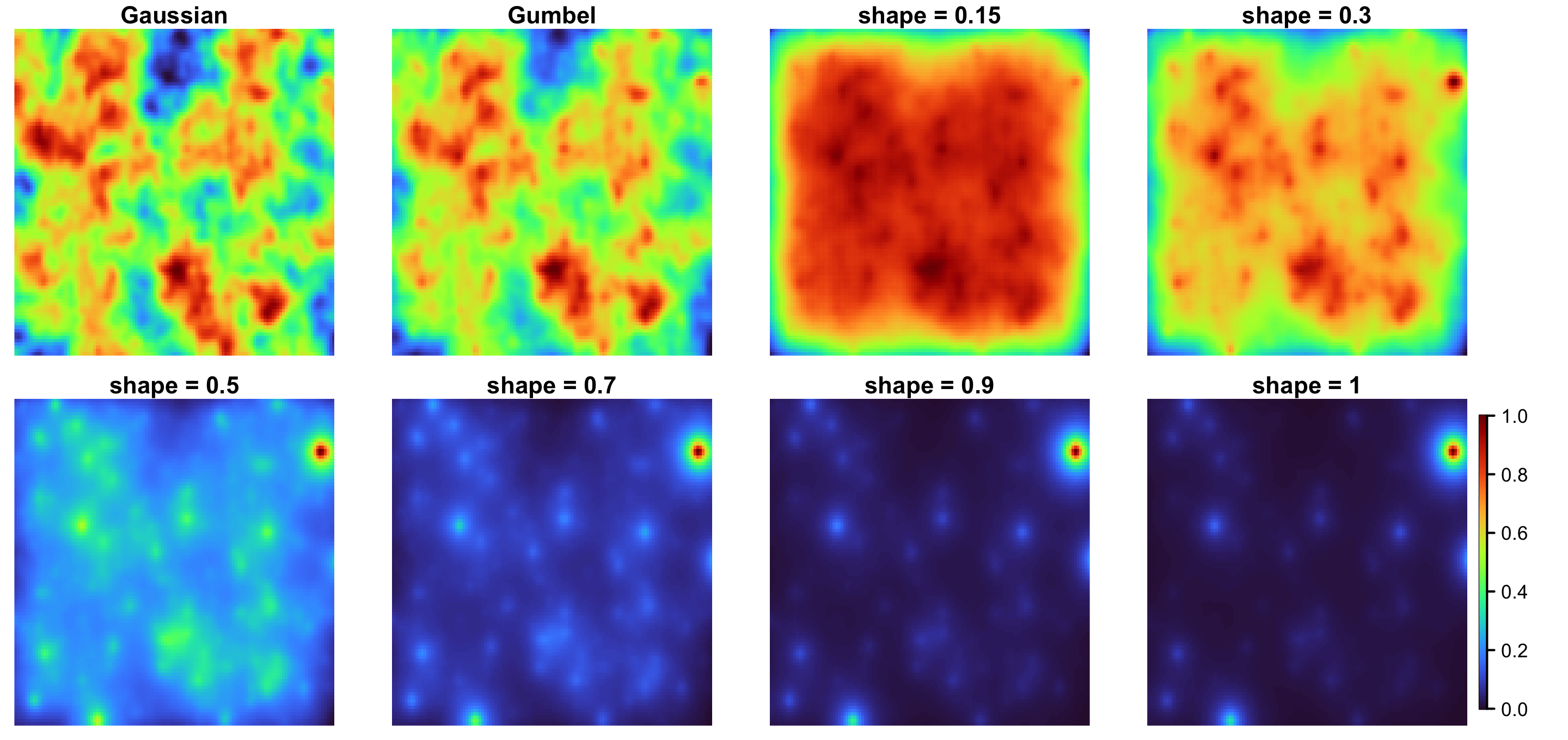}\\
\caption{Expanding a Gaussian SAR process to a Non-Gaussian SAR process using Gumbel\((0,1)\) and GEV\(\bigl(1,\xi,\xi\bigr)\) innovations \(\bigl(\xi>0\bigr)\) 
to generate spatial extreme fields. A fixed uniform sample is used, and then 
transformed for different scenarios, and the resulting spatial fields are 
subsequently scaled to the range \([0,1]\).}
\label{fig:1}
\end{figure}

\noindent \textbf{Model for $g(\bm{s})$.} The spatial process $g(\bm{s})$ is expressed as the sum of the basis functions with associated coefficients. Specifically, we write:
\begin{equation}
\label{multi-model}
g(\bm{s}) = \sum_{j=1}^{m} \varphi_j(\bm{s}) c_j,
\end{equation}
where $\varphi_j(\bm{s})$ are fixed radial basis functions (e.g., Wendland functions \citep{wendland1995, wendland1998}), $m$ represents the number of basis functions, and $c_j$ are random coefficients. 
We further model the coefficient vector as $B\bm{c} = \bm{e}$ where $\bm{e}$ follow a GEV distribution. In this model, the shape parameter $\xi$ of the GEV distribution controls the heavy-tailed behavior of the spatial process, while spatial dependence is incorporated through the SAR matrix $B$, which is tri-diagonal with $B_{j,j} = \kappa^2$, $B_{j,j+1} = -1$, and $B_{j+1,j} = -1$, generalizing a similar construction for Gaussian process modeling \citep{nychka2015multiresolution}, where $\kappa^2$ controls the spatial dependence.  
We assume that $\bm{e} \overset{\text{iid}}{\sim} \text{GEV}(\mu, \sigma, \xi)$, where we set $\mu = 1$ and $\sigma = \xi > 0$, allowing $\xi$ to vary. Fixing $\mu = 1$ and $\sigma = \xi > 0$ ensures that the distribution corresponds to the standard Fréchet form, with the distribution standardized to have location 0 and scale 1 after transformation. Figure~\ref{fig:1} shows the transformation of a spatial field from a Gaussian process to a non-Gaussian spatial extremes field with fixed $\kappa^2=0.1$, under varying GEV shape parameters. As illustrated in Figure~\ref{fig:1}, the spatial field becomes increasingly extreme as the shape parameter increases from $\xi = 0$ to $\xi = 1$, with significant changes in spatial dependence patterns compared to the Gaussian case. It is also important to note that, for a given set of parameters, the simulation of these fields is highly efficient due to the strategic use of sparsity in the key matrices involved in the construction.

Our approach shares conceptual similarities with the hierarchical spatial modeling framework proposed by \cite{reich2012hierarchical, zhang2023flexible}, which supports the feasibility of our framework. The key difference is their assumption of a max-stable process with a fixed marginal distribution, whereas our model differs in that the underlying distribution remains uncertain. Specifically, \cite{reich2012hierarchical} propose a spatial residual variability model that combines a basis expansion with a spatial nugget effect. They assume 
$
y(\bm{s}) \sim \text{GEV}(\mu(\bm{s}), \sigma(\bm{s}), \xi(\bm{s})),
$ 
where $\mu(\bm{s})$, $\sigma(\bm{s})$, and $\xi(\bm{s})$ represent site-specific location, scale, and shape parameters, respectively. The process is then defined as
$$
y(\bm{s}) = \mu(\bm{s}) + \frac{\sigma(\bm{s})}{\xi(\bm{s})} \left(R(\bm{s})^{\xi(\bm{s})} - 1\right),
$$
with $R(\bm{s})$ representing a residual max-stable process following $\mathrm{GEV}(1,1,1)$. This residual process is further specified as $R(\bm{s}) = \zeta(\bm{s})\theta(\bm{s})$, where $\zeta(\bm{s}) \sim \mathrm{GEV}(1, \alpha, \alpha), \alpha \in (0,1)$ captures non-spatial variability, and $\theta(\bm{s})$ is a basis function representation of a positive stable distribution which is closely related to Fr\'echet and has a closed form. While this provides computational advantages, it limits the flexibility of their approach. In contrast, our method does not rely on a max-stable framework or assume a specific marginal distribution.

\noindent \textbf{Presence of Nugget Effect.} In real-world applications, particularly in geosciences, measurements often contain noise, and failing to account for this error can result in misleading inferences and unrealistic models. Defining measurement error in spatial extremes modeling is not straightforward - particularly because incorporating the nugget term makes the likelihood more complex. Additionally, selecting an appropriate distribution for the nugget term poses its challenges. In the model from (\ref{obs-model-eq}), we define a nugget term with a mean 1 and on a log scale, a distribution that closely resembles the Gaussian form. To achieve this, we model the nugget term using the lognormal distribution. Specifically, we assume $\varepsilon \sim \text{lognormal}(\mu', {\sigma'}^2)$,
where \(\tau^2\) controls the nugget variability. The mean of the nugget term is fixed at 1 and its variance to \(\tau^2\) by defining $\mu' = -{\sigma'}^2/{2}$ and ${\sigma'}^2 = \log (1 + \tau^2)$.

\begin{figure}[ht]
    \centering
    \hspace{-1em}
    \includegraphics[height = 2in, width=0.8\textwidth]{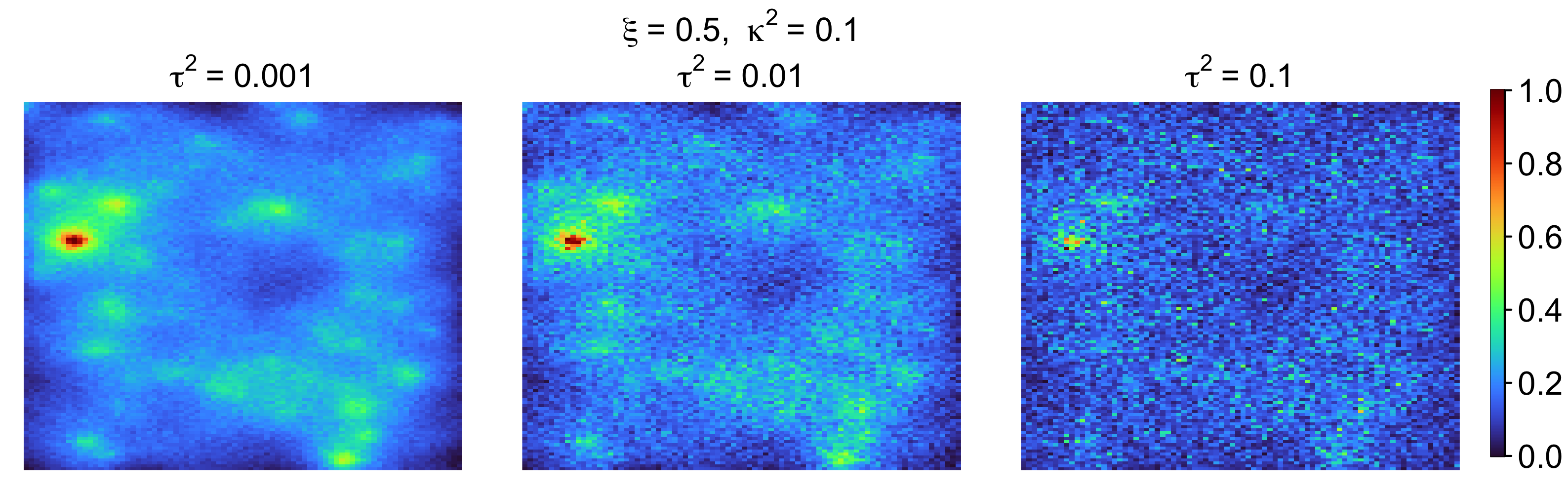}
    \caption{Spatial fields under fixed \(\xi = 0.5\) and \(\kappa^2 = 0.1\) with varying nugget variance \(\tau^2\), ranging from $0.01\%$ to $10\%$ of the total variability. This highlights the influence of the nugget term on the spatial field.}
    \label{fig:SARII}
\end{figure}

To illustrate the effect of the nugget term in the model and justify the choice of its distribution, we fixed the shape and spatial dependence parameters, $\xi$ and $\kappa^2$, in the $g(\bm{s})$ representation and varied the nugget parameter $\tau^2$ to generate $\varepsilon$'s. In Figure~\ref{fig:SARII}, we present the simulated spatial fields $y=g(\bm{s})\cdot \varepsilon$ for fixed $\xi=0.5$ and $\kappa^2=0.1$, with $\tau^2$ varying from 0.01\% to 10\%, to show the influence of the nugget term. 

\section{Parameter Estimation using Convolutional Neural Network (CNN)}
\label{sec:Framework}
Although the model described in (\ref{obs-model-eq}) offers substantial flexibility and computational efficiency for simulating large, non-Gaussian fields, it also introduces several inference-related challenges. In particular, the resulting likelihood becomes intractable, especially with the inclusion of the nugget term, making traditional MLE methods infeasible. Moreover, due to the non-traditional nature of the proposed model only individual components of the model can be interpreted directly. To overcome the computational challenges for making inference, machine learning-based inference methods provide promising alternatives. In particular, recently CNNs have shown considerable success in feature extraction and real-time modeling of spatial data, particularly in applications such as image recognition and spatial feature extraction \citep{liu2018feature}. Following that idea, we propose using a CNN for inference, which takes spatial fields as input and returns estimated values of the model parameters \(\theta = (\xi, \kappa^2, \tau^2) \in \Theta\), where \(\Theta\) is the parameter space, offering a computationally efficient solution for handling large datasets. Next, we describe the CNN framework in greater detail.

\subsection{Details of the Network}
\label{subsec:network}
Typically, a CNN \( \mathcal{N} \) applies a sequence of non-linear transformations over the layers, which map the input spatial field to the output parameter vector. Let \(\bm{y} \) denote the spatial field with dimensions \( d \times d \), and let \( r \) represent the number of replications of the spatial field. The CNN acts as a function \( \mathcal{N} \) which maps the spatial field \( \bm{y} \) to the parameter vector \( \theta \), such that the estimated parameters \( \widehat{\theta} = (\widehat{\xi}, \widehat{\kappa}^2, \widehat{\tau}^2) \) are derived as the corresponding output:
$$
\mathcal{N} : \mathbb{R}^{d \times d \times r} \to \Theta, \quad \text{where} \quad \mathcal{N}(\bm{y}) = \widehat{\theta}.
$$
In this work, we combine dense layers, which fully connect inputs to outputs, with convolutional layers that apply multiple linear filters with a fixed kernel size (e.g., \( M \times M \)) to the input field. A dense layer with \( r_I \) inputs generates \( r_O \) outputs, involving \( O(r_I \times r_O) \) weights, while each convolutional layer has \( O(K \times M^2) \) weights, where \( K\) denotes the number of filters and \( M \) is the kernel size. We apply ReLU/LeakyReLU activation functions after each layer to introduce non-linearity, allowing the model to learn complex relationships between the spatial data and the parameters. The model is trained by optimizing a loss function, which involves adjusting the weights and biases of the network to minimize the discrepancy between the predicted and actual parameter values. This setup allows the optimization to take place locally across small batches and gradually reach a global solution.

Selecting an appropriate loss function is essential for evaluating the accuracy of the CNN. In our approach, we use the mean absolute error (MAE) loss, calculated over a batch of training data, to evaluate the accuracy of the predicted parameters. To guarantee that all parameters in the vector \( \theta = (\xi, \kappa^2, \tau^2) \) contribute equally to the optimization process, we normalize each training parameter by its respective mean and standard deviation as follows.
\begin{eqnarray*}
\Xi_i = \frac{\xi_i - \overline{\xi}_{\text{train}}}{\mathrm{sd}(\xi_{\text{train}})},\ 
\log K^2_i = 
\frac{\log(\kappa^2_i) - \overline{\log(\kappa^2)}_{\text{train}}}
{\mathrm{sd}\!\bigl(\log(\kappa^2)_{\text{train}}\bigr)},\ 
\log \mathcal{T}^2_i = 
\frac{\log(\tau^2_i) - \overline{\log(\tau^2)}_{\text{train}}}
{\mathrm{sd}\!\bigl(\log(\tau^2)_{\text{train}}\bigr)},
\end{eqnarray*}
where \( \overline{(\cdot)}_{\text{train}} \) represents the sample mean of the training parameter \( (\cdot) \), and \( \text{sd}(\cdot)_{\text{train}} \) denotes its sample standard deviation. The MAE loss of 
 $n_B$ samples is defined as:
\begin{eqnarray*}
\mathrm{MAE}(\omega) &=& \dfrac{1}{n_B} \sum_{i=1}^{n_B} \|\theta_i - \widehat{\theta}_i(\omega)\|_1\\
&=& \dfrac{1}{n_B} \sum_{i=1}^{n_B} \left\{ |\Xi_i - \widehat{\Xi}_i(\omega)| + |\log K^2_i - \widehat{\log K^2}_i(\omega)| + |\log \mathcal{T}^2_i - \widehat{\log \mathcal{T}^2}_i(\omega)| \right\},
\label{eq:mae}
\end{eqnarray*}
where \( n_B \) is the batch size, \( \omega \) represents the weights and biases of the network for a given iteration, and \( \|\cdot\|_1 \) denotes the \( \ell_1 \) norm. Minimizing the MAE loss allows the estimates of $\theta$ to closely approximate the true parameter values. Alternative loss functions, such as Huber loss or quantile loss, can also be considered depending on the characteristics of the estimation problem.

\subsection{Network Training}
\label{subsec:train}
To train our neural estimator \(\mathcal{N}\), we explore a wide range of parameter values for \(\theta = (\xi, \kappa^2, \tau^2)\). We generate simulated spatial fields on a regular grid \(\bm{s} = [0, 16] \times [0, 16]\), which is the input size of \(\mathcal{N}\). For each parameter configuration, the innovation vector \(\bm{e}\) is generated using the GEV distribution with the shape parameter \(\xi\), and the SAR matrix \(B\) is computed using \(\kappa^2\) to obtain the coefficients \(\bm{c}\) using the rule \(\bm{c} = B^{-1}\bm{e}\). The coefficients are then fed into the basis representation (defined over \(\bm{s}\)) to construct the initial spatial field. To model the basis functions, we use compactly supported Wendland functions \citep{wendland1995, wendland1998} on a grid of \(16 \times 16\), with an additional \(4 + 4\) functions to account for edge effects during the spatial field simulation. Finally, a nugget term determined by \(\tau^2\) is added, as described in equation (\ref{obs-model-eq}). For all of these computations, we rely heavily on the R packages \texttt{extRemes} \citep{gilleland2015package} for generating the GEV innovations and \texttt{LatticeKrig} \citep{nychka2019package} for generating \(B\) and the basis functions.

\begin{figure}[ht]
    \centering
    \includegraphics[width=1.05\linewidth]{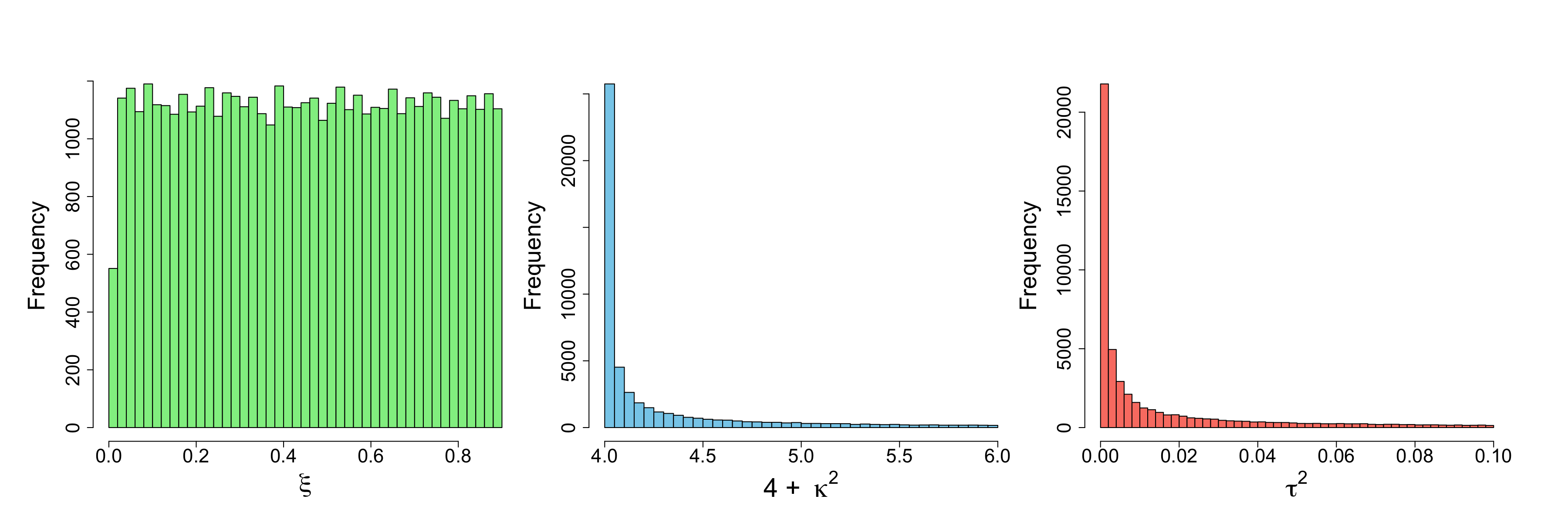}
    \caption{Parameter configurations are generated for training and validating the CNN model. The spatial parameters $\kappa^2$ and $\tau^2$ are generated on the log scale.}
    \label{fig:parameterConfig}
\end{figure}

Finally, we create 50,000 parameter configurations for training and validation, sampling \(\xi \in (0.01, 0.9)\), \(\kappa^2 \in (0.001, 2)\), and \(\tau^2 \in (0.0001, 0.1)\) uniformly, as shown in Figure~\ref{fig:parameterConfig}. 
The ranges for \(\xi\) and \(\kappa^2\) are guided by our initial study, which shows that \(\xi\) close to 1 introduces an extreme scenario, and \(\kappa^2\) close to 2 implies weak spatial dependence. The nugget variability \(\tau^2\) spans 0.01\% to 10\% of the spatial-field noise, as we are working in a multiplicative setup. Requiring \(\xi \in (0.01, 0.9)\) guarantees that the simulated fields exhibit a range of extremes, from none to more pronounced as \(\xi\) approaches 1 (see Figure~\ref{fig:1}).

Once the parameter configuration is defined, we generate the spatial field of size $16\times16\times\text{replication size}$, and further standardize each field by subtracting its median and dividing by its standard deviation (SD). The median and SD are used as they yield better performance in terms of optimization. Next, we train \(\mathcal{N}\) on the Google Colab platform, using 2 virtual CPUs, 32 GB of RAM, and either a P100 GPU (16GB memory) or a T4 GPU (16GB memory, expandable to 25GB of system RAM). Training takes about 90 seconds per epoch, a total of 30 minutes for 100 epochs.
\begin{table}[ht!]
\centering
\begin{tabular}{|l|l|l|l|}
\hline
\textbf{\textit{Layer Type}} & \textbf{\textit{Output Shape}} & \textbf{\textit{Activation}} & \textbf{\textit{Parameters}} \\ \hline
Input Layer & (16, 16, 30) & - & 0 \\ \hline
Conv2D & (14, 14, 64) & LeakyReLU & 17,344 \\ \hline
Conv2D & (12, 12, 32) & LeakyReLU & 18,464 \\ \hline
GlobalAveragePooling2D & (128) & - & 0 \\ \hline
Dense & (512) & ReLU & 16,896 \\ \hline
Dense & (10) & ReLU & 5,130 \\ \hline
Dense & (3) & Linear & 33 \\ \hline
\end{tabular}
\vspace{0.2cm}
\smallskip

\caption{Summary of the CNN model. Total trainable parameters: 58,251. The kernel size is \((3,3)\) for both convolutional layers.  The input layer has shape \((16, 16, 30)\), and the network outputs \((\xi, \log(\kappa^2), \log(\tau^2))\).}
\label{table:1}
\end{table}
The network architecture of \(\mathcal{N}\) is summarized in Table~\ref{table:1}. Instead of using a very dense network, in this work, we opt for a simple architecture to do the inference. To train \(\mathcal{N}\), we adopt the MAE loss and the ADAM optimizer \citep{bock2019proof}, which is well suited for heavy-tailed data. The optimizer updates the weights each time it processes a batch of 100 samples, starting with a learning rate (lr) of $0.001$. LeakyReLU is applied in the convolutional layers to allow small negative gradients, while ReLU in the dense layers ensures computational efficiency and stable gradients. To stabilize training and improve performance, we implement the ``reduce lr on plateau'' callback, which automatically lowers the lr by a factor of 0.1 when the validation MAE  stops improving, thereby allowing more robust optimization. The output layer of \(\mathcal{N}\) employs a linear activation function, producing three scalar values corresponding to the estimated parameters $\xi$, log($\kappa^2$), and log($\tau^2$).

\section{Simulation Study}
\label{sec:sim}
This section presents the results of simulation studies to evaluate the accuracy of the neural estimator $\mathcal{N}$. The evaluation is based on different test sets sampled from  parameter ranges defined in Section~\ref{subsec:train}.

\subsection{Model Evaluation on the Test Set}
In this study, we evaluate the performance of  $\mathcal{N}$ using a test set consisting of $10,000$ distinct parameter configurations, sampled across the defined parameter ranges. The CNN models were trained under three different scenarios, each with a varying number of replications: \( r = 1 \), \( r = 10 \), and \( r = 30 \). 
\begin{figure}[ht!]
    \centering
    \includegraphics[width=\linewidth]{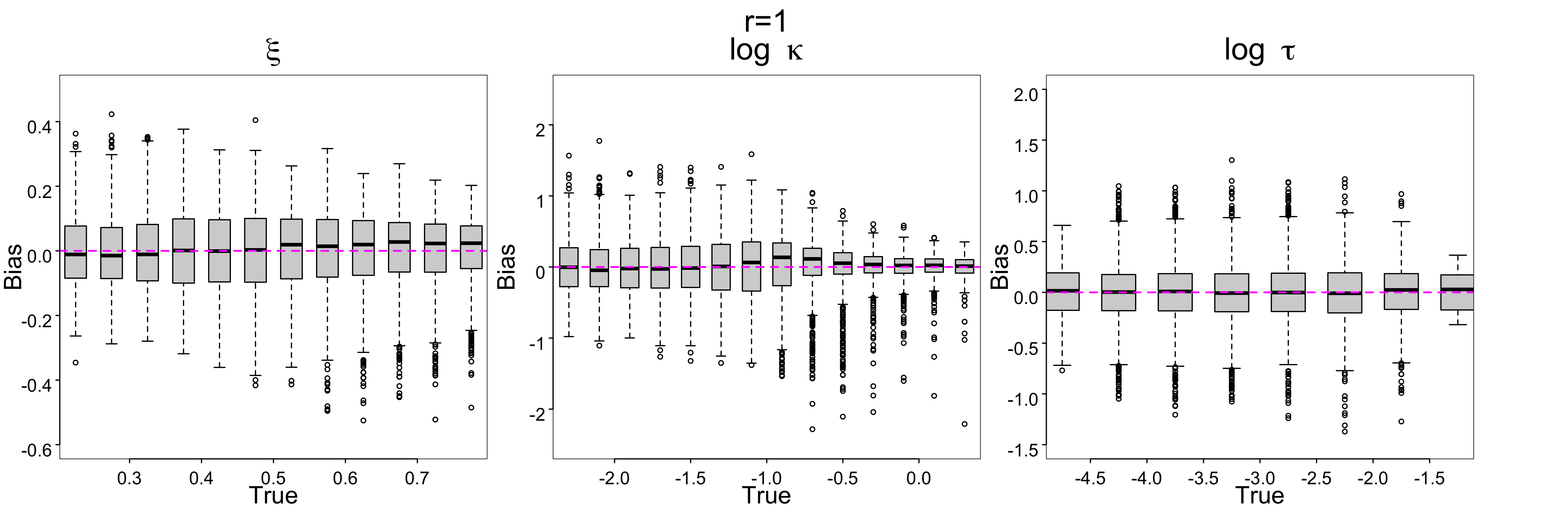}\\
    \vspace{1em}
    \includegraphics[width=\linewidth]{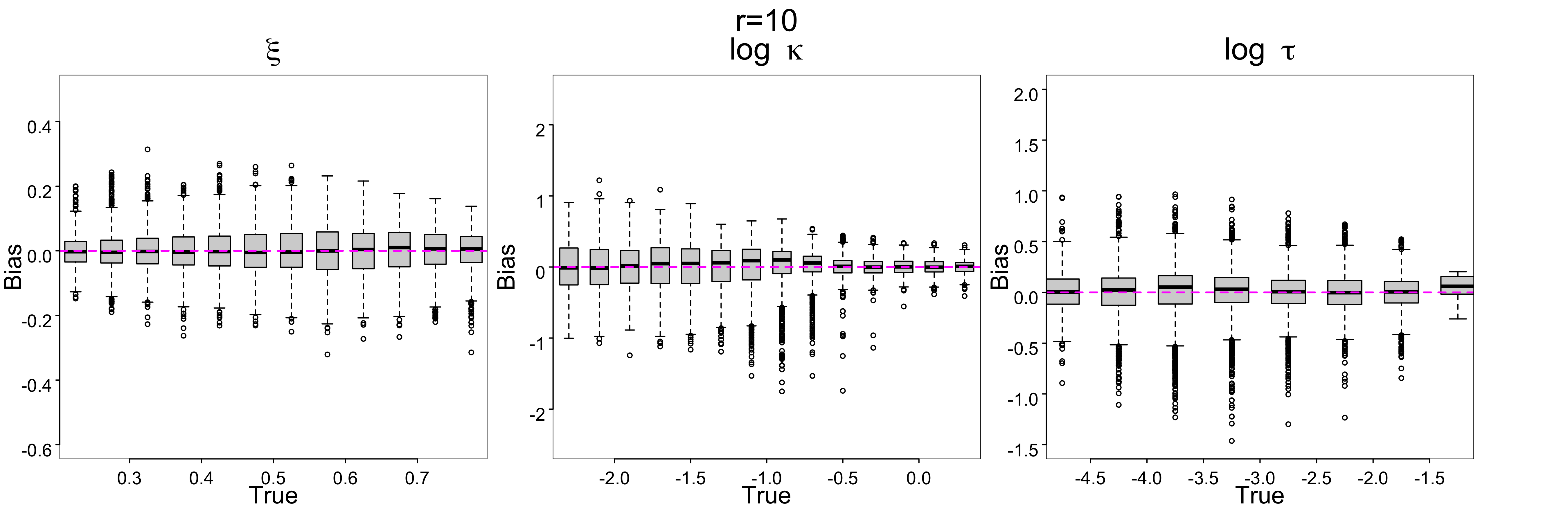}\\ 
    \vspace{1em}
    \includegraphics[width=\linewidth]{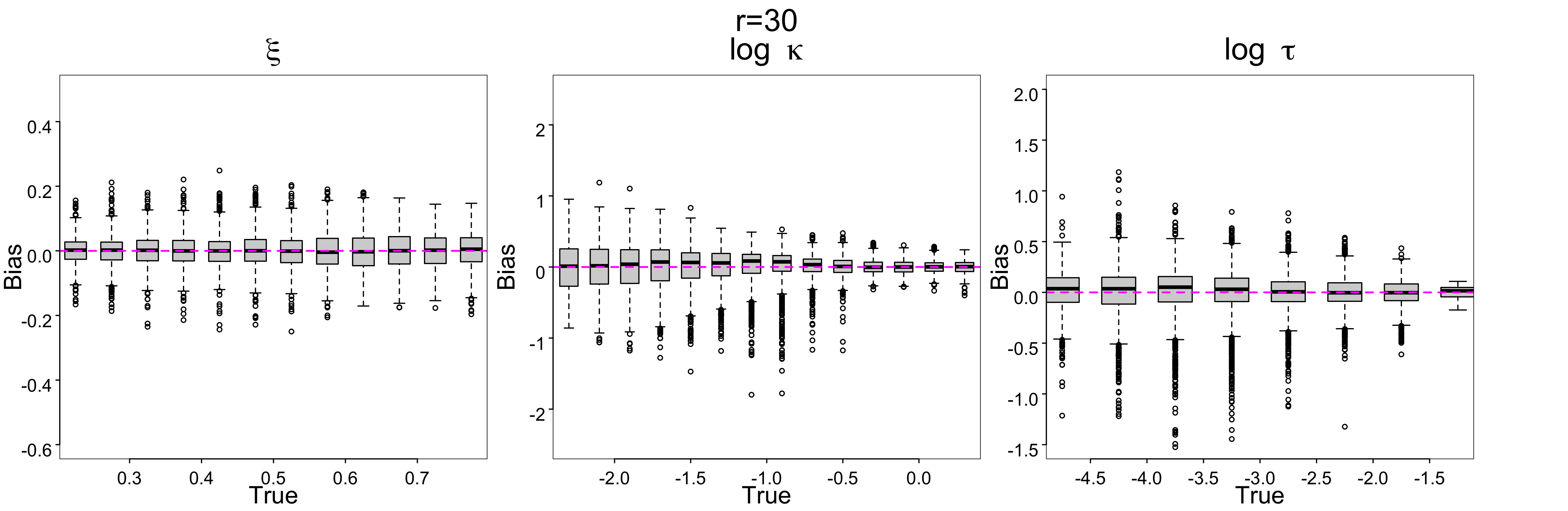}\\ 
    \caption{Performance of the CNN model on the test set of size 10,000, illustrating the bias in parameter estimates for different sample sizes.} 
    \label{fig:TestI}
\end{figure}

While the CNN architecture remains fixed, the input size of the model varies as \( 16 \times 16 \times r \), where \( r \) represents the replicates of the spatial field for a given parameter configuration. These replicates are fed into the network as spatial field samples, entered as channels in the CNN architecture. Next, we evaluate the trained model on the generated test samples with varying replication. The reason for doing the comparison across different replications is to check the behavior of the estimator, as one expects the resulting estimator to exhibit less bias and variability as the replication size increases. The results are visualized in Figure~\ref{fig:TestI}, where the true parameter values are plotted on the $x$-axis, and the $y$-axis represents the bias in the parameter estimates produced by the CNN. The rows in Figure~\ref{fig:TestI} correspond to different replication sizes. As \( r \) increases from 1 to 30, we observe a clear reduction in bias and improvement in the variability, which is expected.

\begin{figure}[ht!]
    \centering
    \includegraphics[width=\linewidth]{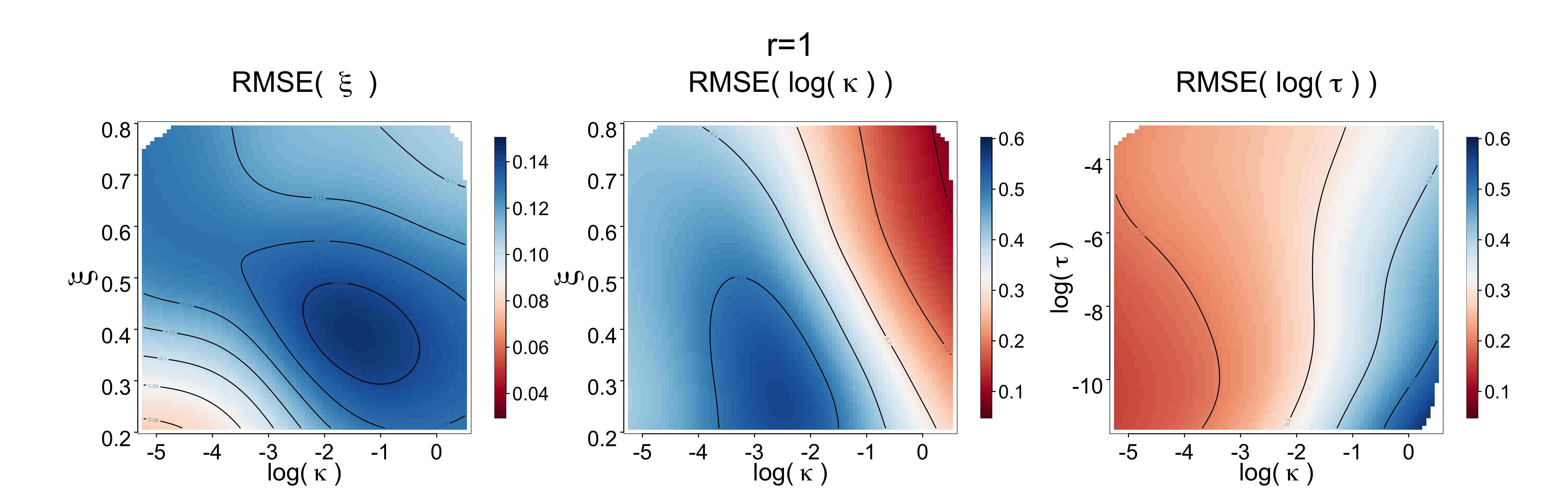}\\
    \vspace{0.8em}
    \includegraphics[width=\linewidth]{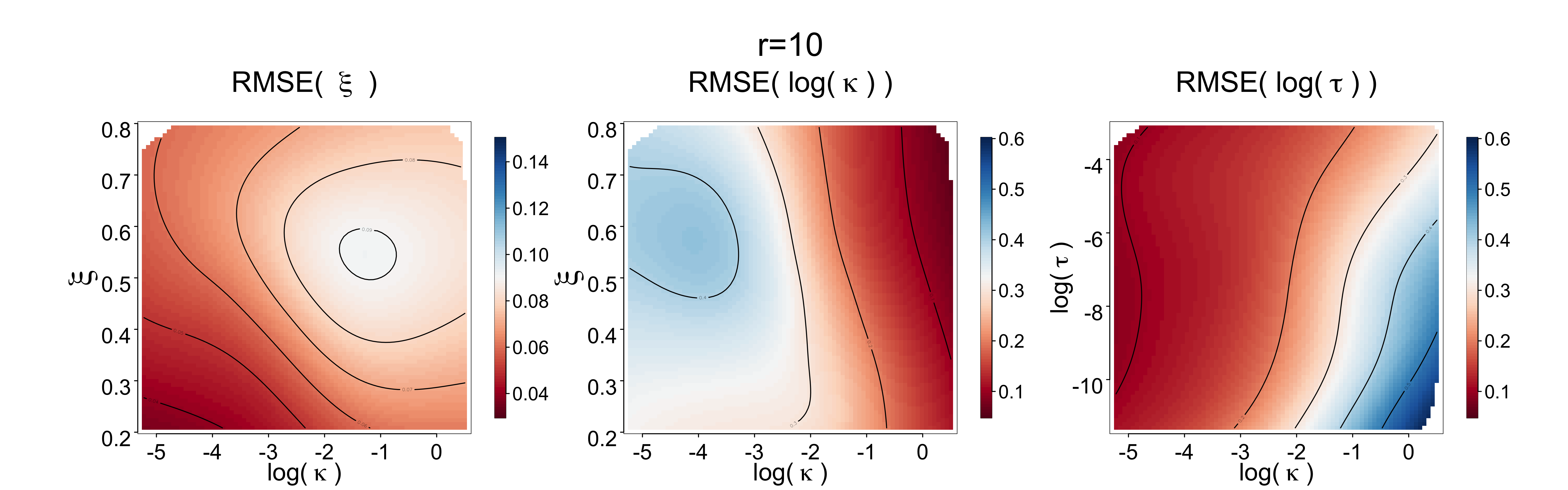}\\ 
    \vspace{0.8em}
    \includegraphics[width=\linewidth]{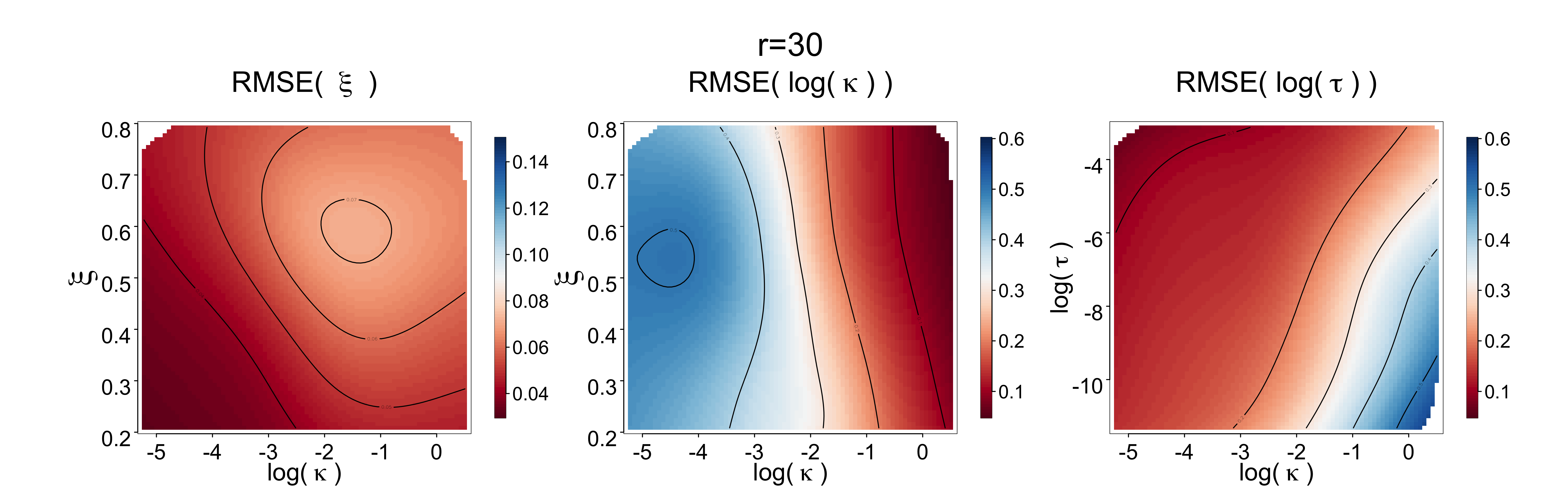}\\ 
    \caption{RMSE of the CNN estimates on the test set with repetition, illustrating the bias and variability in parameter estimates across different replication sizes.}
    \label{fig:TestII}
\end{figure}

To further validate, we compute the root mean squared error (RMSE) across the varying replication sizes. To do this, we generate a smaller test set with repetition to calculate both bias and variability in the inference. By repetition, we mean repeating the simulation process for the same parameter configuration multiple times, including simulating the spatial field with replications. Specifically, the test set has 1,000 unique parameter configurations, each repeated 60 times. Figure~\ref{fig:TestII} presents the RMSE for these parameters across different replication sizes. We observe clear improvements in estimation accuracy as the replication size increases.

\subsection{Comparison with Maximum Likelihood Estimates}
In this section, to compare the CNN-based estimates with MLEs, we consider a simplified setting that excludes the nugget term. As noted earlier, the proposed model in (\ref{obs-model-eq}) yields an intractable likelihood function when the nugget term is included. In no-nugget-effect scenarios, typically observed in satellite data and in climate model outputs derived from global emulators, the term \(\varepsilon(\bm{s})\) in (\ref{obs-model-eq}) is omitted. Consequently, the model simplifies to
\begin{eqnarray}
y(\bm{s}) = g(\bm{s}) 
= \sum_{j=1}^m \varphi_j(\bm{s}) \, c_j,
\label{eqn:no-nugget-effect}
\end{eqnarray}
where \(\bm{c} = (c_1, \ldots, c_m)\) can be written as
$$
\bm{c} = B^{-1}\bm{e}, 
\quad
\bm{e} = (e_1, \ldots, e_m),
\quad
e_i \overset{\text{iid}}\sim \text{GEV}(1, \xi, \xi)
\quad i=1,\ldots,m.
$$
The likelihood of the model described above in (\ref{eqn:no-nugget-effect}) is tractable, but optimizing it to calculate the MLE is time-consuming and often challenging. Therefore, the following study not only compares the behavior of MLEs and CNN-based estimators, but also aims to provide an alternative and significantly more computationally efficient method for parameter estimation, specifically in the absence of the nugget effect in the model.
\begin{figure}[ht!]
\centering
\includegraphics[width=\linewidth]{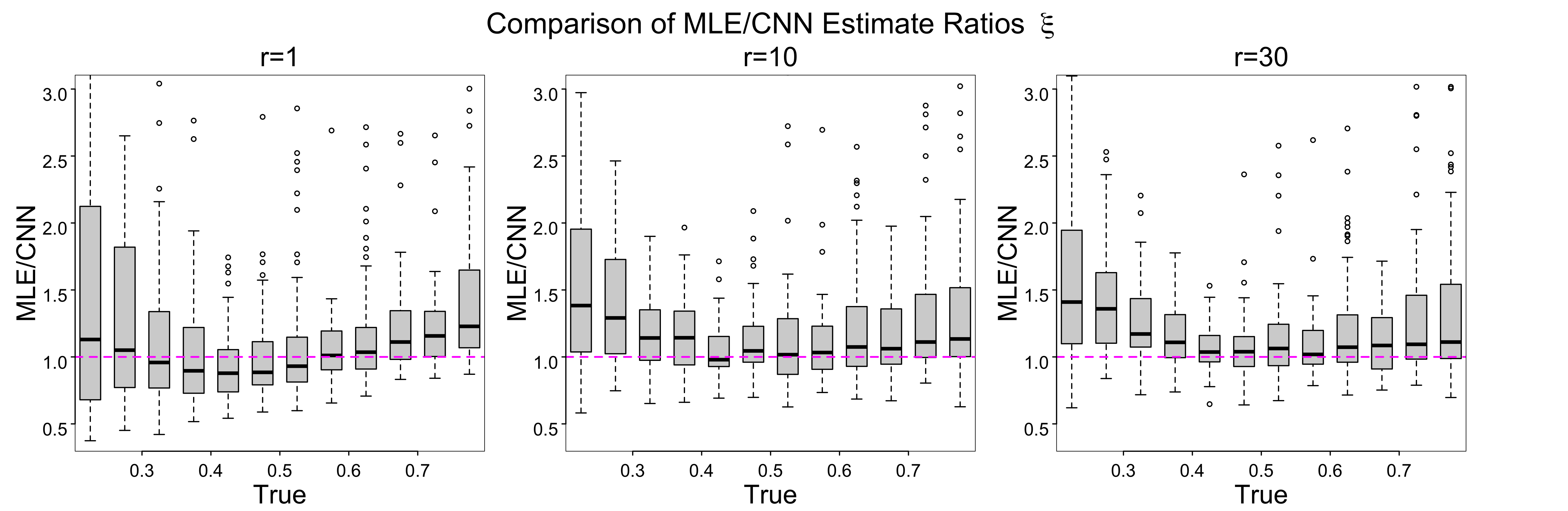}\\
\vspace{1.2em}
\includegraphics[width=\linewidth]{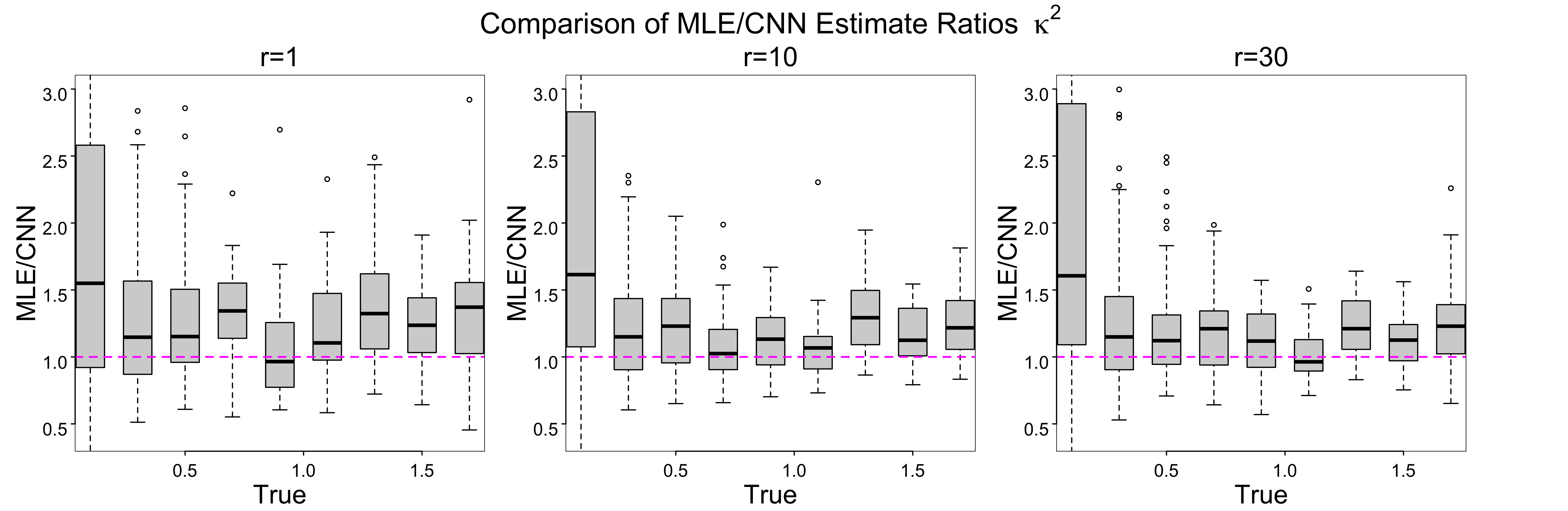}\\
\vspace{1em}
\caption{Comparing CNN estimates  with MLEs in the no-nugget-effect scenario, shown as the ratio \(\text{MLE}/\text{CNN}\) for various true parameters. Different sample sizes are shown.}
\label{fig:TestIV}
\end{figure}

\begin{figure}[ht!]
\centering
\includegraphics[width=0.9\linewidth]{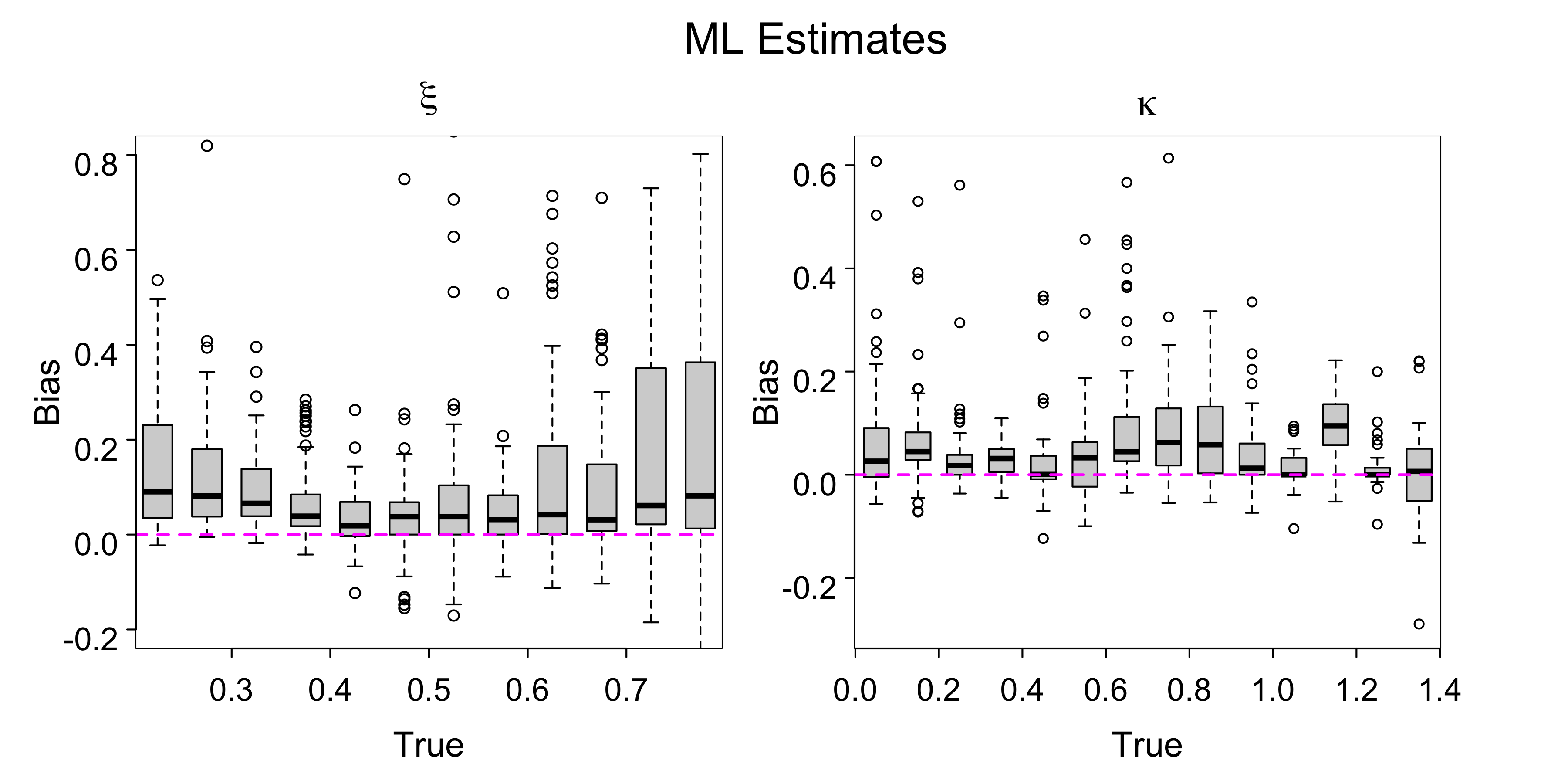}\\ 
\includegraphics[width=0.9\linewidth]{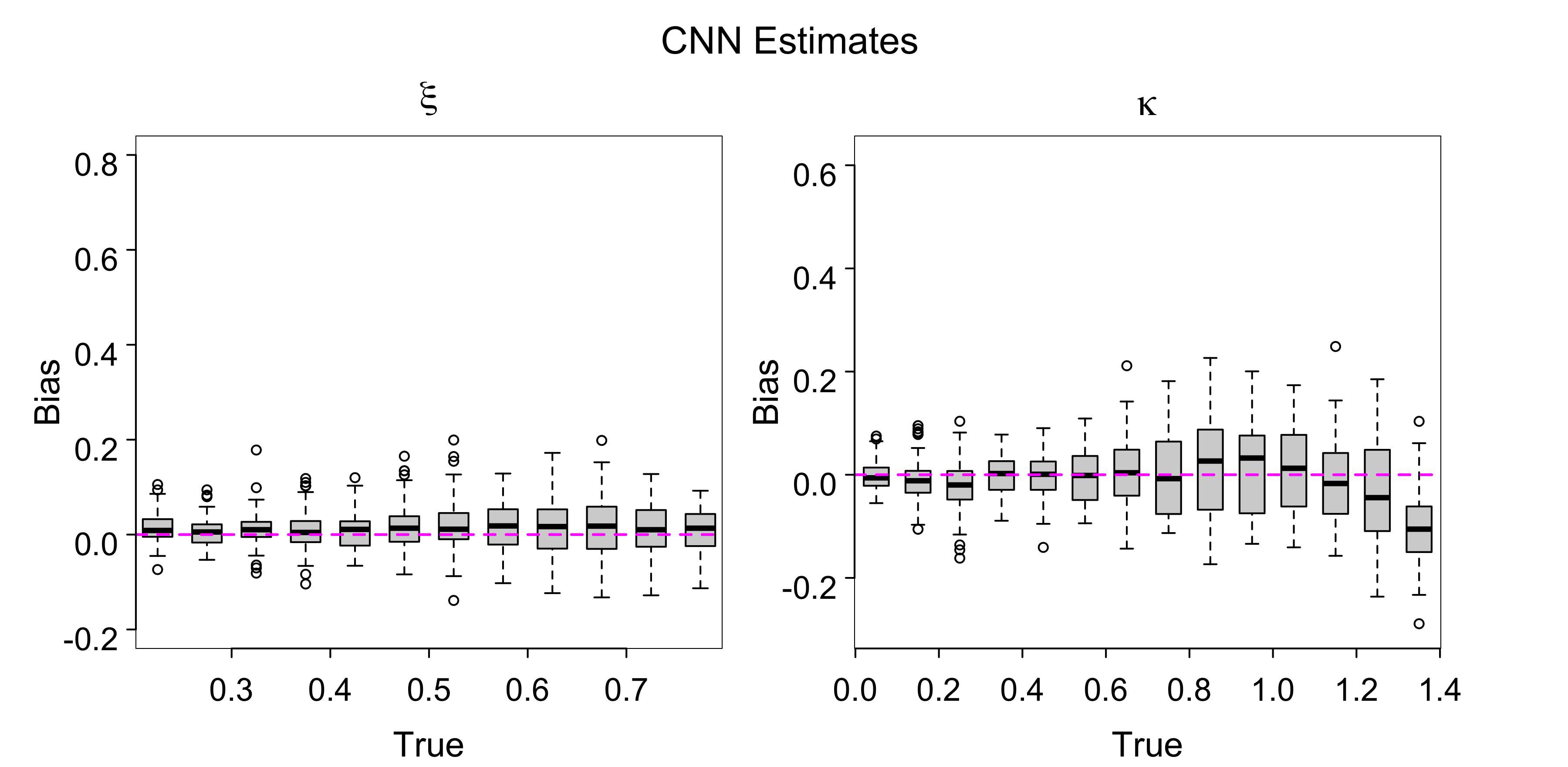}\\ 
\caption{Bias plots of the MLE (top) versus the true parameter, and the CNN estimates (bottom). Both plots correspond to the test sample for the 30-replication case.}
\label{fig:TestIII}
\end{figure}

For this study, we construct a test set of size \(1,024\) by defining a \(32 \times 32\) grid of \((\xi, \kappa^2)\) values, drawn from the parameter ranges outlined in Section~\ref{subsec:train}.
We generate spatial fields of dimension \(16 \times 16 \times r\), where \(r = 1, 10, 30\), and for each test set across varying replications, we compute both MLEs and CNN estimates. Figure~\ref{fig:TestIV} presents the ratio of MLE to CNN estimates across different sample sizes. When using the ratio to compare the estimates, variation in the ratio indicates systematic overestimation or underestimation by one estimator relative to the other. Figure~\ref{fig:TestIII} shows the bias of the MLE relative to the true parameters, alongside the bias of the CNN estimates. Based on Figures~\ref{fig:TestIV}-\ref{fig:TestIII}, the CNN appears to outperform the MLE in terms of parameter estimation.

The MLEs are computed using the \texttt{optim} function in \texttt{R} with the Nelder-Mead method, running on a laptop equipped with an 8-core M3 chip and 16GB of RAM. The likelihood optimization takes approximately 5.3 hours to process the entire test set. In contrast, evaluating the trained CNN on the same test set takes only about 7 milliseconds, demonstrating a substantial computational advantage.

\subsection{Uncertainty Quantification on Test Samples} 
\label{sec:uq}
To quantify the uncertainty in parameter estimates generated by $\mathcal{N}$, we propose using quantile regression (QR) \citep{koenker2001quantile}. In QR, the model fits the conditional quantile function rather than the conditional mean, as in ordinary least squares. In particular, we fit quantile regression at the $2.5^{\text{th}}$ and $97.5^{\text{th}}$ percentiles to provide the lower and upper bounds of the $95\%$ confidence interval. The parameter configurations used during training and validation of the neural network are used to fit the quantile curves. 

Our model consists of three parameters, denoted by $\theta=$ $(\xi,\kappa^2, \tau^2)$. After training $\mathcal{N}$, we evaluate the training data to obtain the estimates $\widehat \theta =(\widehat\xi, \widehat\kappa^2, \widehat\tau^2)$. These estimates from $\mathcal{N}$ are then used as covariates in the QR model, with the corresponding true parameters serving as the response variables: 
\begin{eqnarray*}
    \xi\sim(\widehat\xi,\text{log}(\widehat\kappa),\text{log}(\widehat\tau)),\ \text{log}(\kappa) \sim(\widehat\xi,\text{log}(\widehat\kappa),\text{log}(\widehat\tau)), \hspace{0.5em}\text{and}\ \text{log}(\tau) \sim(\widehat\xi,\text{log}(\widehat\kappa),\text{log}(\widehat\tau)).
\end{eqnarray*}
Note that, incorporating all three estimated parameters into the QR fit accounts for the additional variability introduced by the other estimates, improving the confidence bounds and coverage of each individual parameter. For the spatial dependence and nugget variability parameters, we use the logarithm of their values in the formulation to mitigate additional variability.

\begin{figure}[ht!]
    \centering
    \hspace{-2em}
    \includegraphics[width=1.02\linewidth]{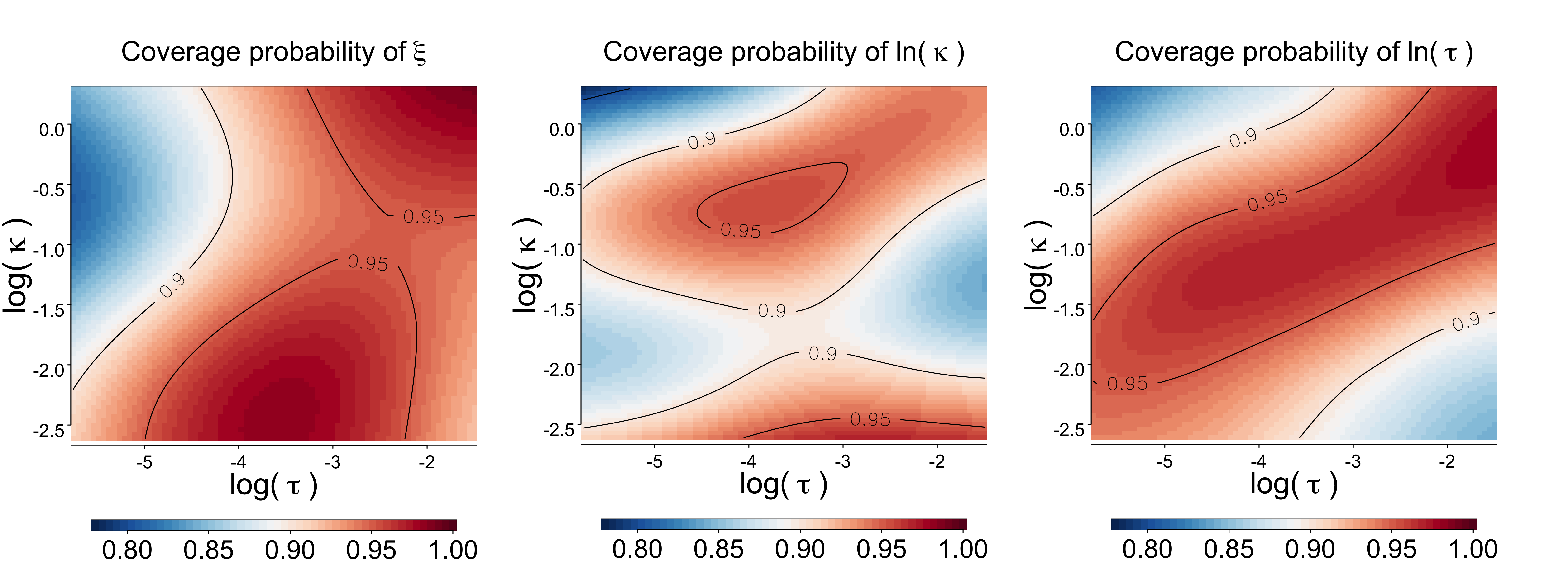}
    \caption{Surface probability plot for the 95\% coverage of the CNN estimates across the true log($\kappa$) and true log($\tau$).}
    \label{fig:Coverage}
\end{figure}

For the computation, we define a test sample of size \(343 = 7 \times 7 \times 7\), constructed over a parameter grid with \(\xi \in (0.2, 0.8)\), \(\kappa^2 \in (0.005, 1.8)\), and \(\tau^2 \in (0.0001, 0.05)\). For each configuration on this grid, we generate 500 repetitions to compute the 95\% confidence bounds. This setup results in \(16 \times 16 \times 30\) spatial fields per parameter configuration, and the process is repeated 500 times across various configurations to generate samples for calculating the 95\% coverage probability. Figure~\ref{fig:Coverage} presents the coverage probabilities across the true \(\log(\kappa)\) and \(\log(\tau)\) for the test sample.
Overall, the coverage aligns with the 95\% level (denoted by the contour lines), although it can drop to about 80\% in some regions, mainly due to the confounding effect or sampling error. We plot the coverage surface of each parameter based on the true \(\log(\kappa)\) and \(\log(\tau)\) because they appear to introduce greater variability estimation and making the confidence interval wider. Despite these variations, most of the coverage probabilities remain close to \(0.90\) to \(0.95\), indicating that our QR-based uncertainty quantification adequately reflects the uncertainty in parameter estimates.

\section{Application to RCM Precipitation Output}
\label{sec:CS}
For further illustration, we implement our model on precipitation data from the North American Coordinated Regional Climate Downscaling Experiment (NA-CORDEX) \citep{nacordex} Weather Research and Forecasting (WRF) RCM simulations for the period 1980--2010. These simulations were driven by ERA-Interim boundary conditions as part of the NA-CORDEX project. NA-CORDEX provides high-resolution regional climate projections for North America, with the WRF model dynamically downscaling coarser global climate model outputs to a 25-km projected grid. Although the grid is curvilinear, we treat it as regular and rectangular in our analysis.

\begin{figure}[ht!]
    \centering
    \includegraphics[width=\linewidth]{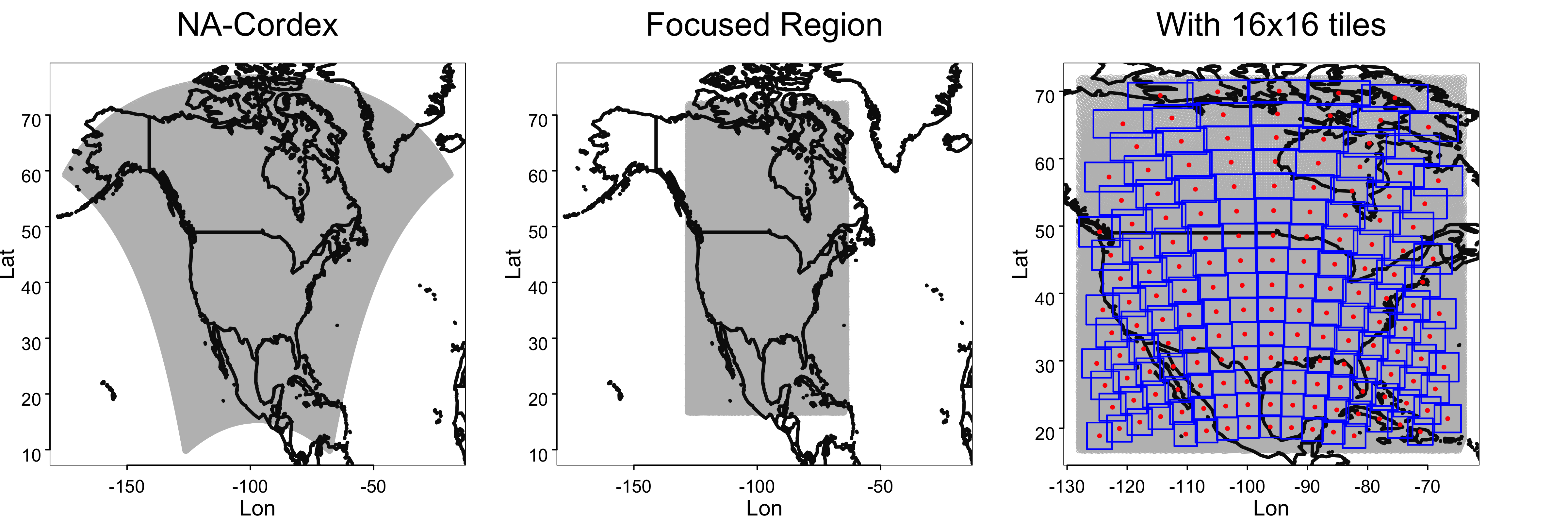}\\
    \includegraphics[width=\linewidth]{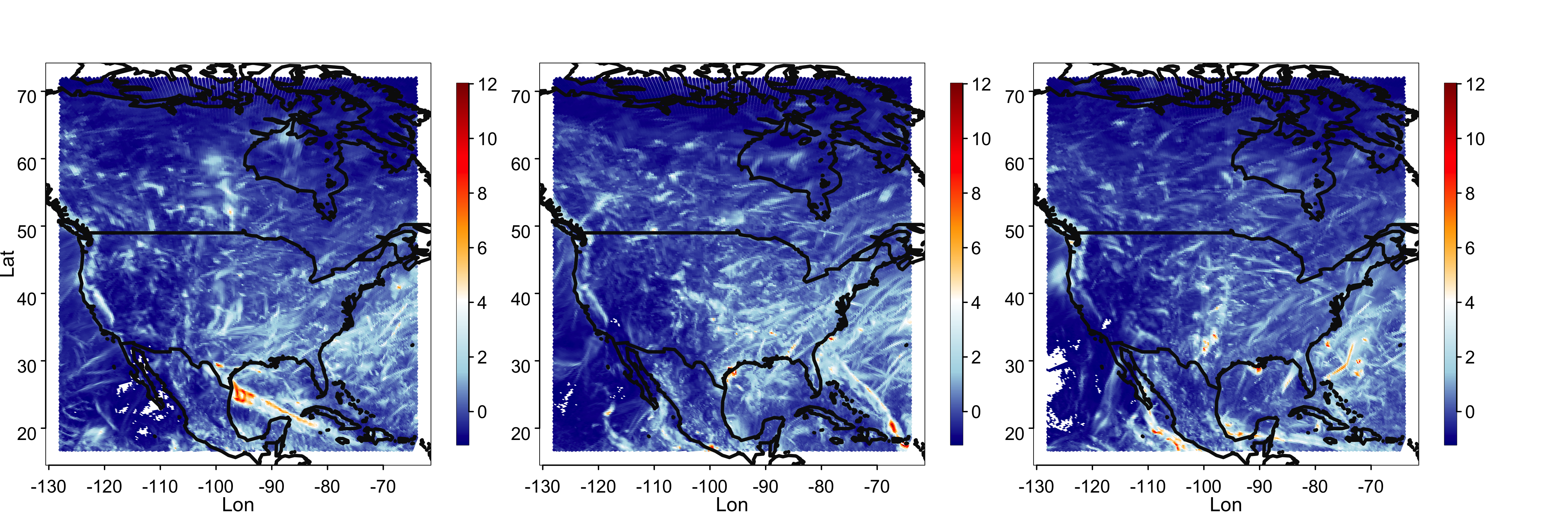}\\
    \caption{We present the NA-CORDEX precipitation observations across the spatial domain for the specified projection over North America, highlighting the focused region and the defined 16$\times$16 windows over the focused region for $\mathcal{N}$ evaluation (top). Additionally, realizations of the spatial process for annual maxima over the focused region are shown for randomly selected years (bottom).}
    \label{fig:CSI}
\end{figure}

The spatial domain consists of a $297\times 281$ grid. For the 31-year precipitation dataset, we focus only on the annual maxima across space for spatial extremal modeling. The spatial domain and the corresponding spatial extreme fields extracted from the annual precipitation maxima for three randomly selected realizations are illustrated in Figure~\ref{fig:CSI}. Clearly, we observe more pronounced extremal behavior near the Gulf Coast and the Pacific West region of the US. Since the spatial domain includes both land and ocean, we prioritize land areas as much as possible. Next, we define $16\times16$ non-overlapping tiles across the focus domain, as shown in Figure~\ref{fig:CSI} to match the input shape of the \(\mathcal{N}\) model and perform estimation on each tile. The resulting parameter estimates are centered within each tile. We further bias corrected the estimates across the tiles to ensure that the values fall within an interpretable range. More specifically, we fit a smooth spline function to the training sample which is then applied to correct the estimated parameters across the tiles for the bias correction.

\begin{figure}[ht!]
    \centering
    \includegraphics[width=\linewidth]{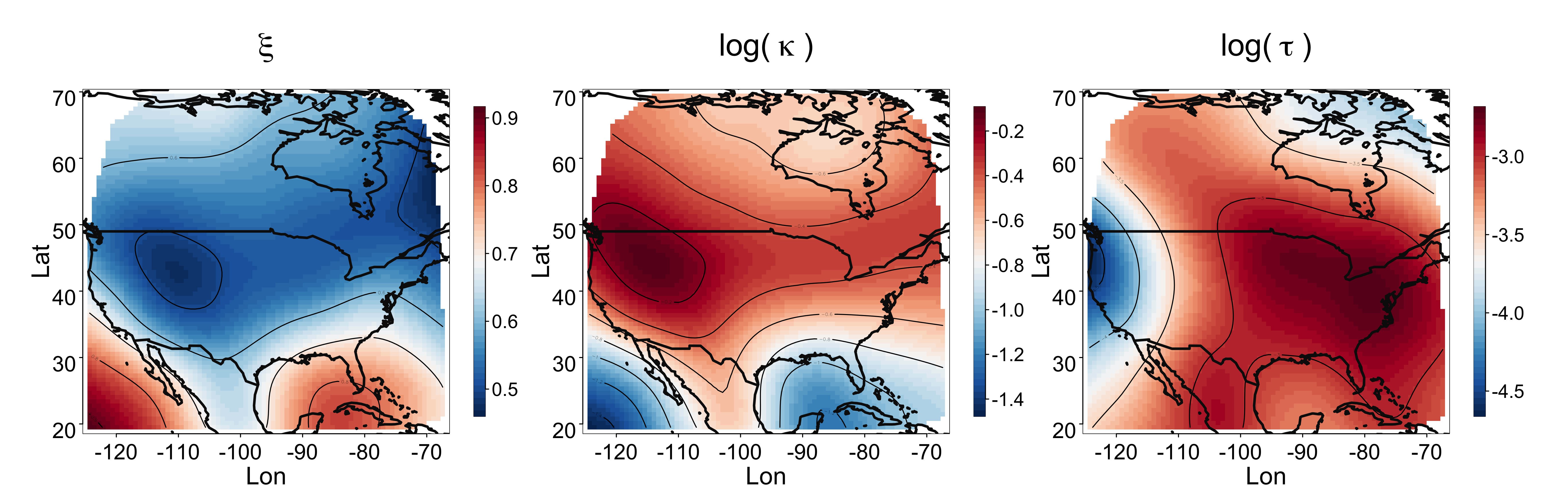}\\
    \caption{Smoothed surface of model parameter estimates across the 16×16 tiles centered on the focused region of the NA-CORDEX annual maximum precipitation observations.}
    \label{fig:CNN-CS}
\end{figure}

We obtain a total of 166 tiles in the focus domain and evaluate the extremal observations using the CNN. In Figure~\ref{fig:CNN-CS}, we present a smoothed surface of model parameter estimates. These parameter estimates show higher shape parameter (\(\xi\)) values near the gulf coast and the pacific west region, suggesting a greater likelihood of extreme precipitation events. This finding aligns with the observed data presented in Figure~\ref{fig:CSI}, validating the model's effectiveness.

We also observe that lower $\xi$ values in the northern US and Canada indicate less frequent and less intense extreme precipitation.

From the plot of estimated spatial dependence parameter ($\kappa^2$), we observe a higher $\kappa^2$ value in the eastern US, indicating lower spatial dependence, suggesting lower-scale precipitation extremes, while the western US exhibits comparatively stronger dependence, likely due to the influence of local convective storms and orographic effects. For the nugget variability ($\tau^2$), we observe higher variability across the eastern and central US, implying more localized variability in extreme precipitation. In contrast, the west coast and mountainous regions (e.g., Rockies) exhibit lower nugget variability, suggesting more structured precipitation patterns.  

\begin{figure}[ht!]
    \centering
    \includegraphics[width=0.66\linewidth]{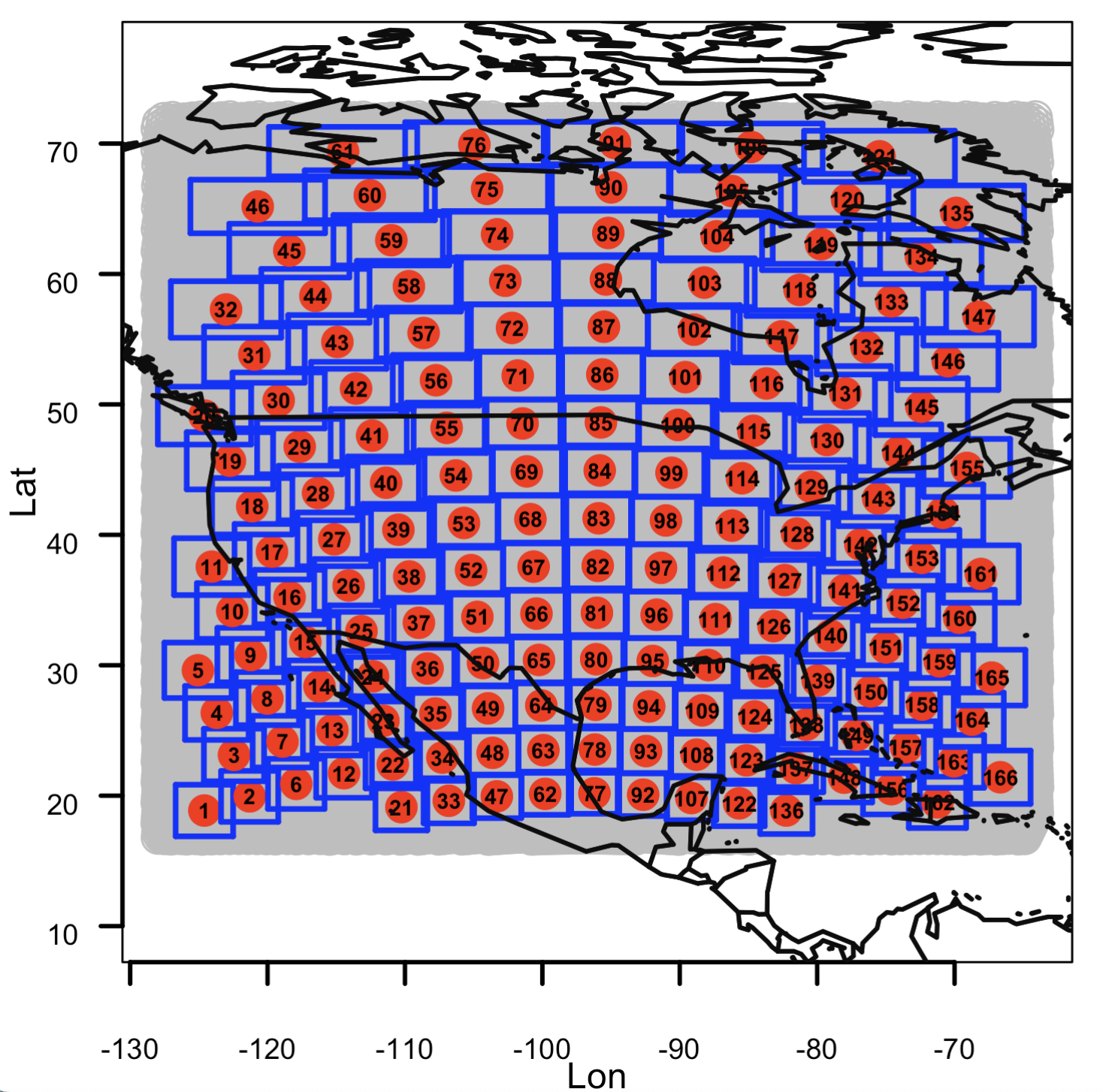}\\
    \caption{Focused NA-Cordex spatial domain, with 16x16 tiles, with numbering.}
    \label{fig:Tiles}
\end{figure}

\begin{figure}[ht!]
    \centering
    \includegraphics[width=0.9\linewidth]{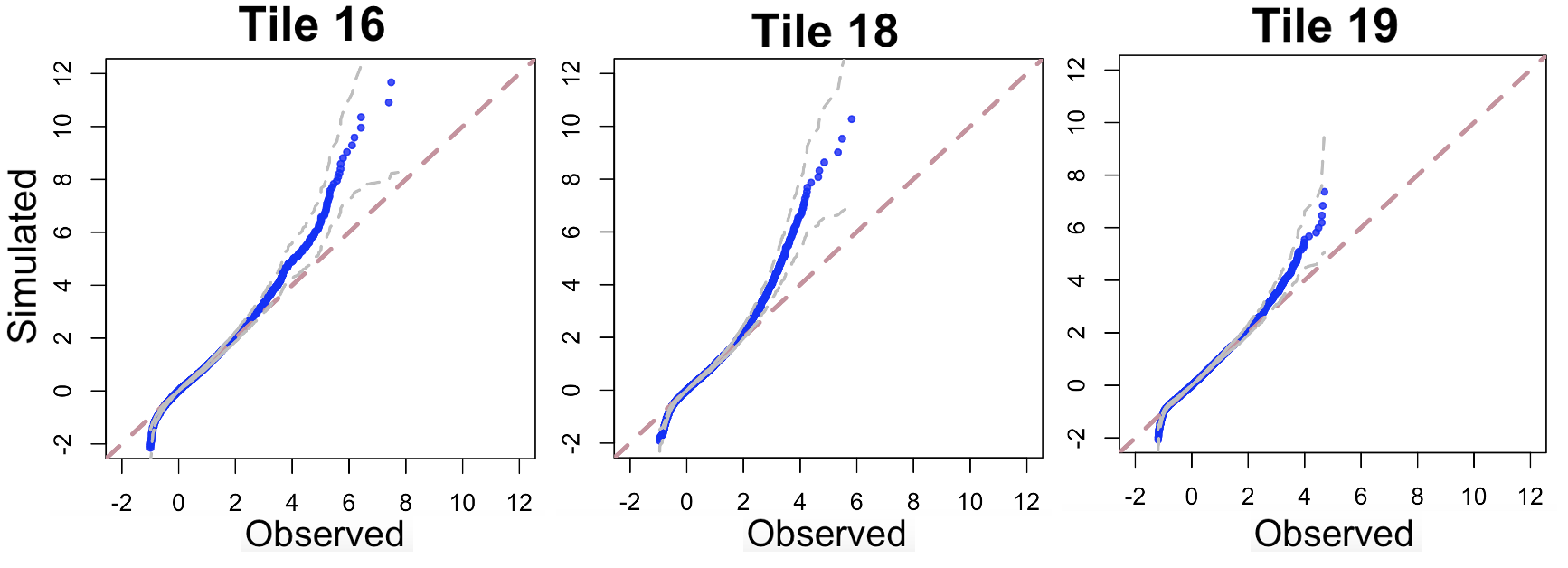}\\
    \vspace{1em}
    \includegraphics[width=0.9\linewidth]{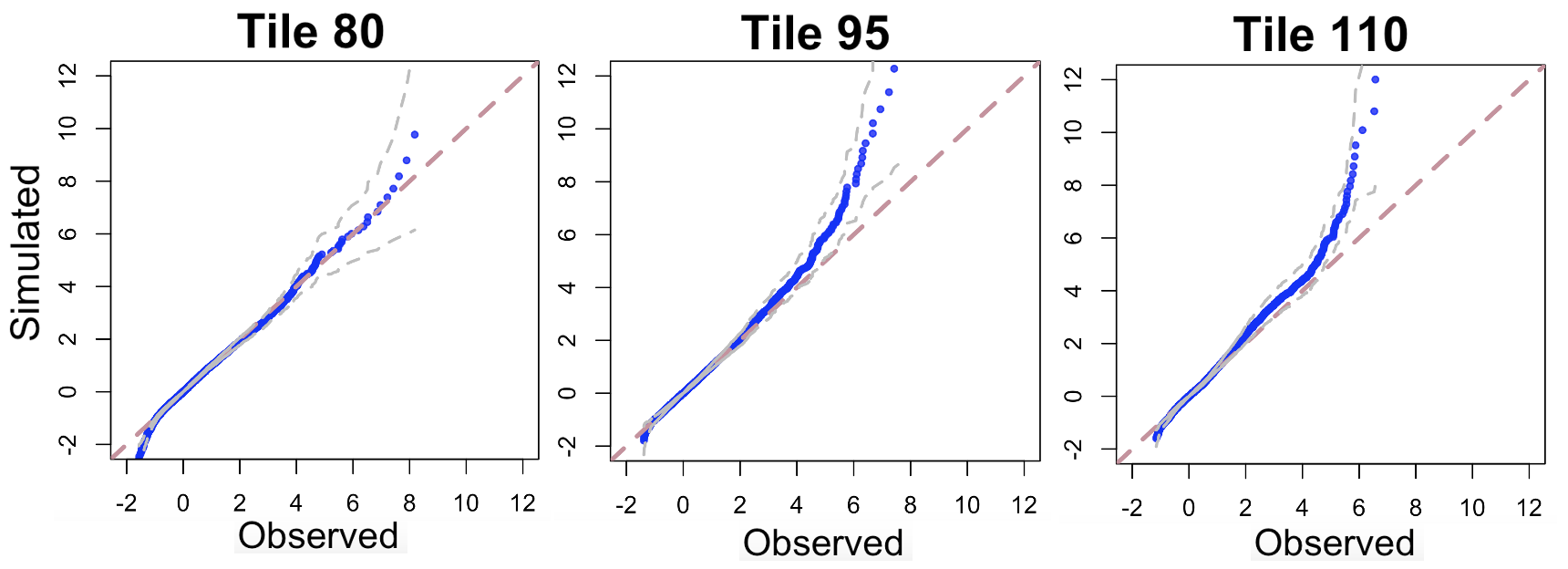}\\
    \vspace{1em}
    \includegraphics[width=0.9\linewidth]{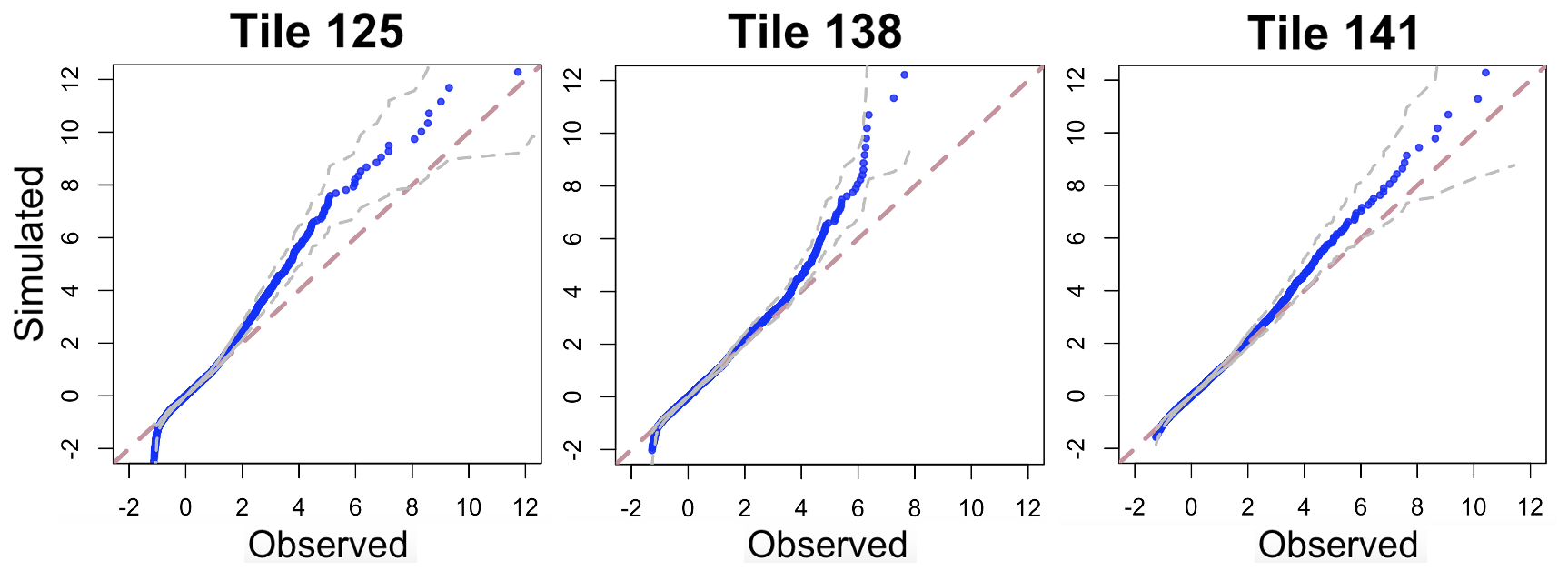}  \\
    \caption{QQ plot of observed vs. simulated fields for randomly selected tiles using CNN estimates. The spatial fields are standardized before comparison. The pink line represents the ideal fit, blue points show the QQ plot, and the grey lines indicate the lower and upper quantile bounds.}
    \label{fig:CS-QQ}
\end{figure}

\begin{figure}[ht!]
    \centering
    \includegraphics[width=0.9\linewidth]{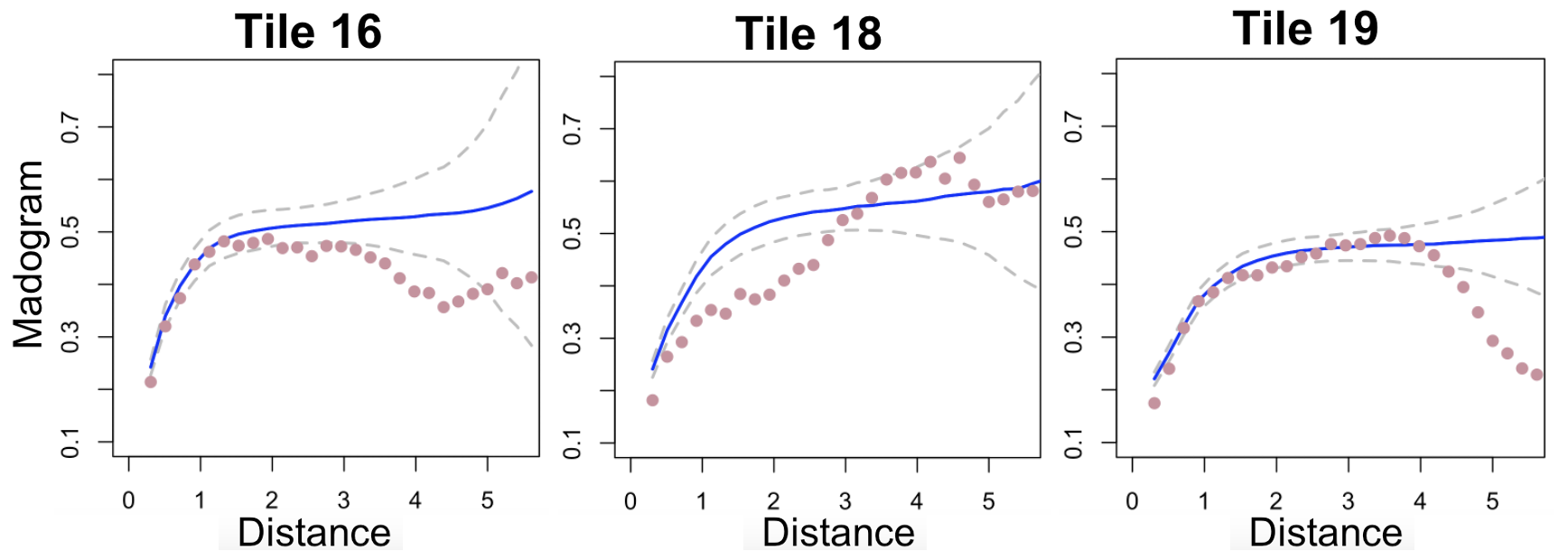}\\
     \vspace{1em} 
    \includegraphics[width=0.9\linewidth]{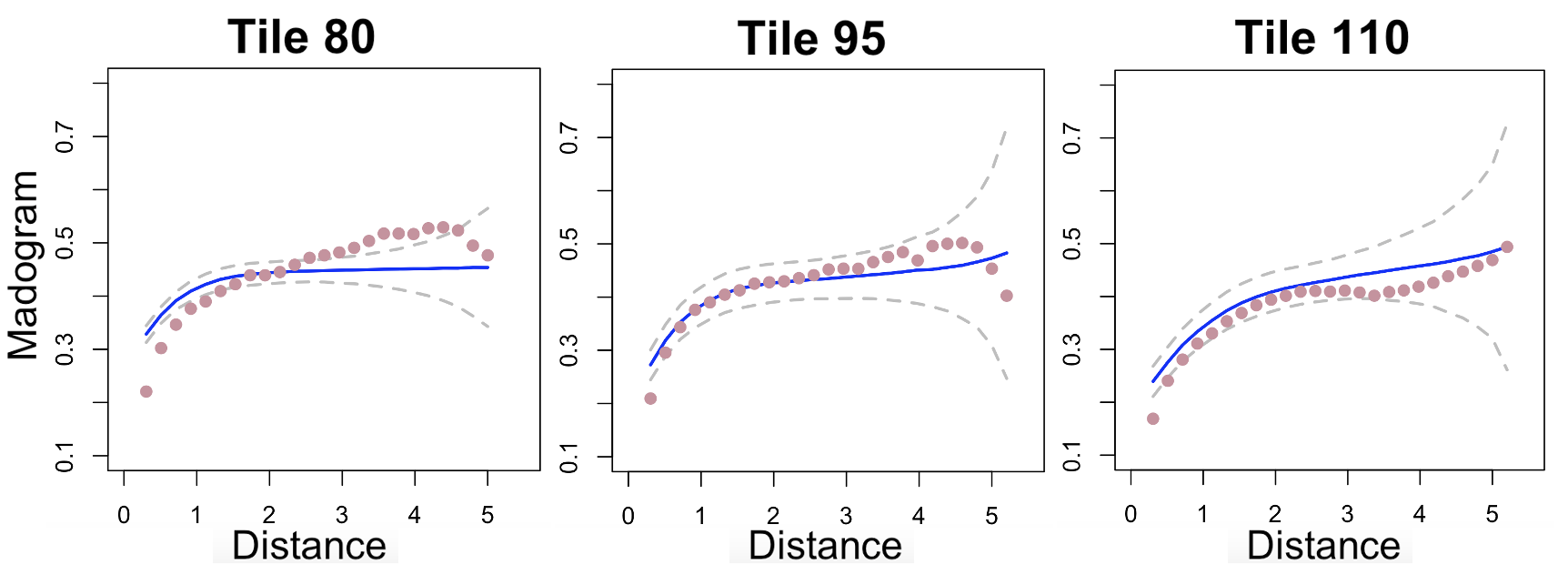}\\
    \vspace{1em}
    \includegraphics[width=0.9\linewidth]{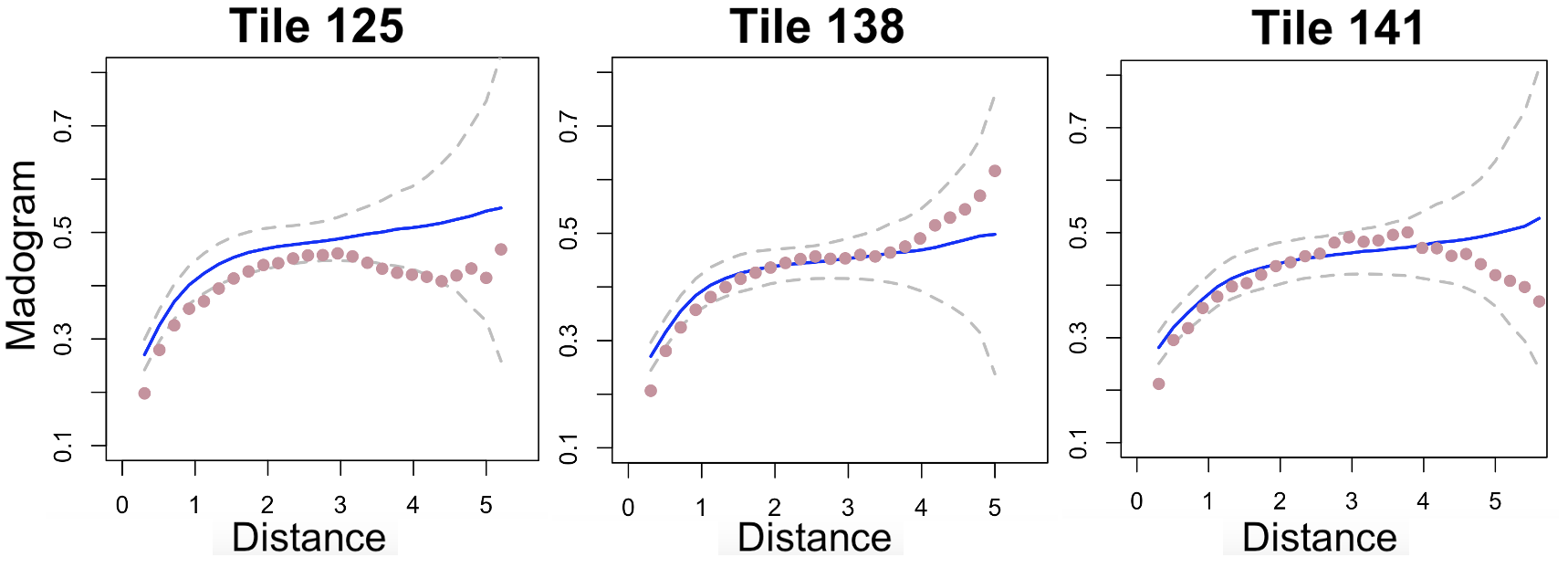}  \\
    \caption{Fitted madogram plot for estimated parameters across randomly selected \(16\times16\) tiles for standardized observed vs. simulated data. The pink dotted line represents the madogram of the observed process, while the blue line shows the median madogram over $400$ repetitions. The grey dotted lines indicate the $95\%$ confidence bounds.}
    \label{fig:Mado}
\end{figure}

For further evaluation, we simulate the process at the tile centers and compare it with the observed process in terms of quantiles. Before generating the quantile-quantile (QQ) plot for further computation, we apply a bias correction to the simulated process using quantile mapping and also standardize both the simulated and observed spatial processes using their median-IQR. To better summarize the results, we present findings from a few selected tiles, primarily focusing on regions in the US that typically experience maximum precipitation: the pacific west and parts of the south-eastern US. The corresponding tile numbers are shown in Figure~\ref{fig:Tiles}. The simulation is performed on these tiles for 400 repetitions over a 30-year period for comparison. As shown in Figure~\ref{fig:CS-QQ}, the simulated fields closely resembles the observed fields, in terms of quantiles. To further assess the similarity, we compute the madogram of the spatial field within the tiles. The madogram is a simple and common tool used in spatial extremes to evaluate spatial dependence as an alternative to the variogram \citep{cooley2006variograms, naveau2009modelling}. The madogram is defined as:
$$
2{\gamma}(\bm{h}) = \mathbb{E} \left| Y(\bm{s}_i) - Y(\bm{s}_j) \right| ,
$$
where \( Y(\bm{s}_i) \) and \( Y(\bm{s}_j) \) are the values of the spatial field at locations \( \bm{s}_i \) and \( \bm{s}_j \), respectively. We compute the empirical madogram using:
$$
2\widehat{{\gamma}}(\bm{h}) = |N(\bm{h})|^{-1} \sum_{(i,j) \in N(\bm{h})} \left| Y(\bm{s}_i) - Y(\bm{s}_j) \right|,
$$
where \( N(\bm{h}) \) is the set of all pairs of locations \( (\bm{s}_i, \bm{s}_j) \) separated by a distance approximately equal to \( \bm{h} \), and \( |N(\bm{h})| \) represents the number of such pairs. Since the madogram is based on absolute differences, it is less sensitive to extreme values. In Figure~\ref{fig:Mado}, we observe that the madogram computed from the observed field aligns well with that of the simulated process, with most values falling within the CI bounds. These results support the effectiveness of the proposed model in capturing spatial extremal behavior and detecting hotspots in the spatial process.


\begin{figure}[ht!]
    \centering
    \includegraphics[width=0.7\linewidth]{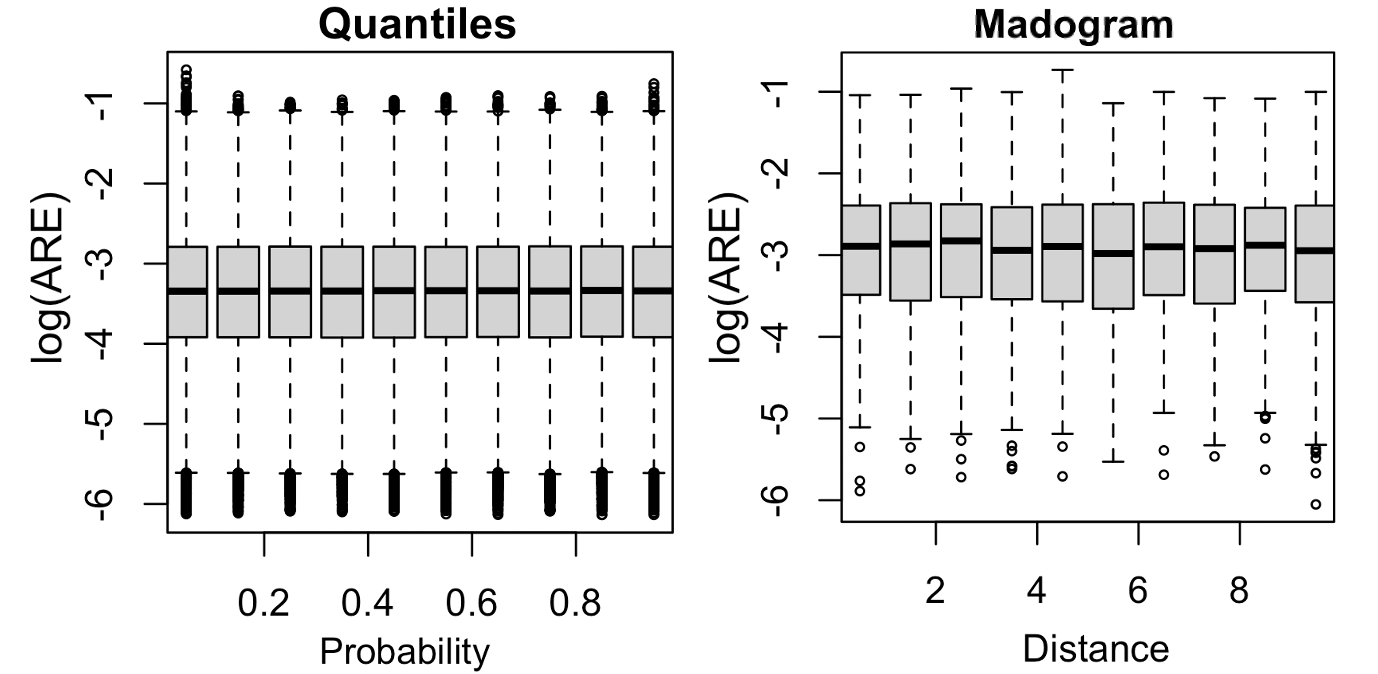}\\  
    \caption{Computing the absolute relative error (ARE) to compare the observed and simulated process behavior in terms of quantiles across different probability levels and madograms across varying distances across all 166 tiles.}
    \label{fig:ARE}
\end{figure}

To summarize the model fit across all defined tiles, we compute the absolute relative error (ARE) for both the observed quantiles and the average quantiles from the simulated data across the probability values used for quantile computation. Similarly, we compute the madogram for the observed process and compare it with the average madogram computed using the simulated data. In Figure~\ref{fig:ARE}, we present the results, showing that the log ARE for both quantile and madogram comparisons falls within the negative range and remains low, indicating a decent fit of the model.

\section{Conclusion}
\label{sec:conc}

In this study, we develop a novel framework for modeling spatial extremes using a non-Gaussian SAR model with GEV innovations. Given the intractable nature of the resulting likelihood, we propose a computationally efficient approach for parameter estimation using CNNs. Specifically, our model is trained on a diverse set of spatial fields, simulated over a broad range of parameter values, ensuring it captures the heavy-tailed and spatially heterogeneous characteristics of extreme events.

Our model is evaluated on different test samples, both generated on a defined grid and randomly selected across the parameter configuration, with and without repetition. We also compare the estimates from the CNN with those from the MLE approach, under the assumption of no nugget effect. Overall, we find that the CNN model handles the estimation process quite effectively. In addition, we apply our method to annual maximum precipitation data from NA-CORDEX WRF simulations, highlighting its ability to extract meaningful spatial patterns. The model is able to identify regions prone to extreme precipitation, with higher shape parameter estimates $\xi$ observed in the gulf coast and pacific west regions, consistent with known climatological patterns. Furthermore, spatial dependence $\kappa^2$ is stronger in the western US, indicating large-scale precipitation extremes, while higher nugget variability $\tau^2$ in the central and eastern regions suggests more localized variations in extreme events.

To quantify uncertainty, we propose using quantile regression, using the same training dataset to fit the lower and upper confidence curve for inference, which provided robust 95\% confidence intervals for parameter estimates. The madogram analysis and quantile-quantile plots further validate the model's ability to capture the extremal structure of observed precipitation fields. In partiuclar, our new method exhibits strong agreement with traditional likelihood-based estimates, while offering a substantial computational advantage - achieving parameter inference in milliseconds compared to hours for maximum likelihood estimation.

Despite its advantages, our approach has some limitations. One key aspect is the inclusion of covariates in spatial modeling, which is crucial for accounting for external factors that may influence extreme events. While this work briefly touches on a potential method to estimate the fixed effects in the model using an iterative backfitting approach, a more comprehensive analysis incorporating covariates is a focus of our future work. This extension will allow for a deeper understanding of how covariates impact spatial extremes and improve the model's applicability in real-world scenarios. Next, while CNNs show considerable promise in extracting spatial patterns, their interpretability remains a challenge compared to traditional statistical models. Additionally, the use of CNNs imposes restrictions based on the input shape, necessitating local likelihood estimation and defining moving tiles for evaluating larger spatial domains. An alternative approach would be to train a CNN on a larger input size, but this could prove computationally expensive in terms of both simulation and evaluation. Furthermore, the performance of the neural network heavily relies on the coverage of the training data, meaning it may require further refinement when applied to new, unseen climate conditions. In addition, we plan to explore alternative architectures, such as transformers and normalizing flows, for improved flexibility in modeling complex spatial dependencies.

This study represents an important first step toward developing a new, flexible, and computationally efficient framework for spatial extremes that completely bypasses the traditional use of max-stable formulations. This is a significant advancement in the field, as it addresses some of the weaknesses inherent in traditional methods. A potential drawback of this framework is the intractable nature of the resulting likelihood. However, with the recent surge in research focused on deep learning methods for spatial fields, we further advance spatial extreme event modeling by proposing the use of CNNs for parameter estimation. In conclusion, by combining statistical rigor with computational efficiency, our framework offers a scalable tool for climate risk assessment, providing a useful decision-making tool in the face of increasing environmental variability.

\section*{Code Availability}
All code used to generate the results presented in this paper is publicly available at the following GitHub repository: \url{https://github.com/Sweta-AMS/GEV-SAR}. The repository includes scripts for constructing and training the neural network model, evaluating its performance on test datasets of varying sample sizes, and generating the associated uncertainty quantification results. In addition, the repository provides the trained neural network model to facilitate replication and further evaluation.

\bibliographystyle{agsm}
\bibliography{bibliography.bib}
\end{document}